%% file: main.tex
\documentclass[conference]{IEEEtran}

\input{configs}

\begin{document}

\title{
\textsc{\interpret}: Interactive Predicate Learning from Language Feedback for Generalizable Task Planning
}

\author{Muzhi Han$^1$, Yifeng Zhu$^2$, Song-Chun Zhu$^1$, Ying Nian Wu$^1$, Yuke Zhu$^2$ \\
$^1$University of California, Los Angeles, $^2$The University of Texas at Austin \\
\textbf{\url{https://interpret-robot.github.io}}
}

\maketitle

\begin{abstract}
Learning abstract state representations and knowledge is crucial for long-horizon robot planning. We present \interpret, an \ac{llm}-powered framework for robots to learn symbolic predicates from language feedback of human non-experts during embodied interaction. The learned predicates provide relational abstractions of the environment state, facilitating the learning of symbolic operators that capture action preconditions and effects. By compiling the learned predicates and operators into a \ac{pddl} domain on-the-fly, \interpret allows effective planning toward arbitrary in-domain goals using a PDDL planner. In both simulated and real-world robot manipulation domains, we demonstrate that \interpret reliably uncovers the key predicates and operators governing the environment dynamics. Although learned from simple training tasks, these predicates and operators exhibit strong generalization to novel tasks with significantly higher complexity. In the most challenging generalization setting, \interpret attains success rates of 73\% in simulation and 40\% in the real world, substantially outperforming baseline methods.

\end{abstract}

\IEEEpeerreviewmaketitle

\section{Introduction}

Effective long-horizon planning is a long-standing challenge in robotics~\cite{zhu2021hierarchical,xu2019regression,zhang2023learning}. Imagine a household robot that prepares a meal in your kitchen. It must be capable of generating faithful multi-step action plans to manipulate novel objects and achieve diverse task goals. Recently, \acfp{llm} have shown the ability to decompose a high-level task goal into semantically meaningful sub-tasks leveraging the vast amount of world knowledge they encode~\cite{li2022pre,huang2022language}. They exhibit the emergent property of acquiring planning capabilities from a few in-context examples~\cite{dong2022survey,huang2022language}. Researchers have successfully applied \ac{llm}-based planners in real-world robotic tasks~\cite{ahn2022can,huang2022inner,singh2023progprompt,liang2023code}, where they can easily incorporate various forms of feedback and produce plans in novel situations. Nevertheless, \ac{llm}-based planners still struggle to generalize strongly to long-horizon tasks, and they offer no performance guarantees~\cite{liu2023llm+,valmeekam2022large,silver2022pddl}.

In contrast, classical planners~\cite{lavalle2006planning,russell2010artificial} based on symbolic abstractions provide complementary strengths in generating long-horizon plans with formal guarantees. At the heart of these planners are \textit{predicates}, which are binary-valued functions that map environment states to high-level symbolic representations, \textit{e.g.}, a function that transforms the workspace observation into semantic relations such as \texttt{on\_table(apple)}. With these symbolic predicates, we can subsequently model state transitions with symbolic \textit{operators}~\cite{fikes1971strips}, describing the preconditions and effects of the robot's actions on the symbolic states. The predicates and operators together form a \ac{pddl} domain~\cite{fox2003pddl2}, enabling a planning algorithm to generate plans for arbitrary in-domain tasks~\cite{helmert2006fast}. Despite the wide adoptions of planning algorithms in robotics~\cite{kaelbling2011hierarchical,toussaint2015logic,garrett2020pddlstream}, these methods usually require substantial manual effort and domain expertise to meticulously design the predicates and operators, hindering their applicability to real-world problems.

To combine the best of both worlds, there has been a growing interest in integrating learning methods with planning algorithms.  Notable efforts have been made to learn symbolic representations from interaction data through unsupervised learning methods~\cite{konidaris2018skills,james2022autonomous,loula2019discovering,silver2023predicate,ahmetoglu2022deepsym}. However, without explicit guidance, they struggle to uncover predicates that capture task-relevant semantic relations to facilitate planning. Meanwhile, cognitive studies~\cite{mandler1992build,goksun2010trading} have shown that human infants are remarkably efficient in acquiring new predicate-like relational concepts, such as spatial relations for stacking blocks, from the language feedback of caregivers during physical play. Inspired by these studies, we envision an \textit{interactive learning} scheme that will enable a robot to rapidly learn useful abstractions for planning from online human feedback.

\begin{figure}[t!]
    \centering
    \includegraphics[width=\linewidth]{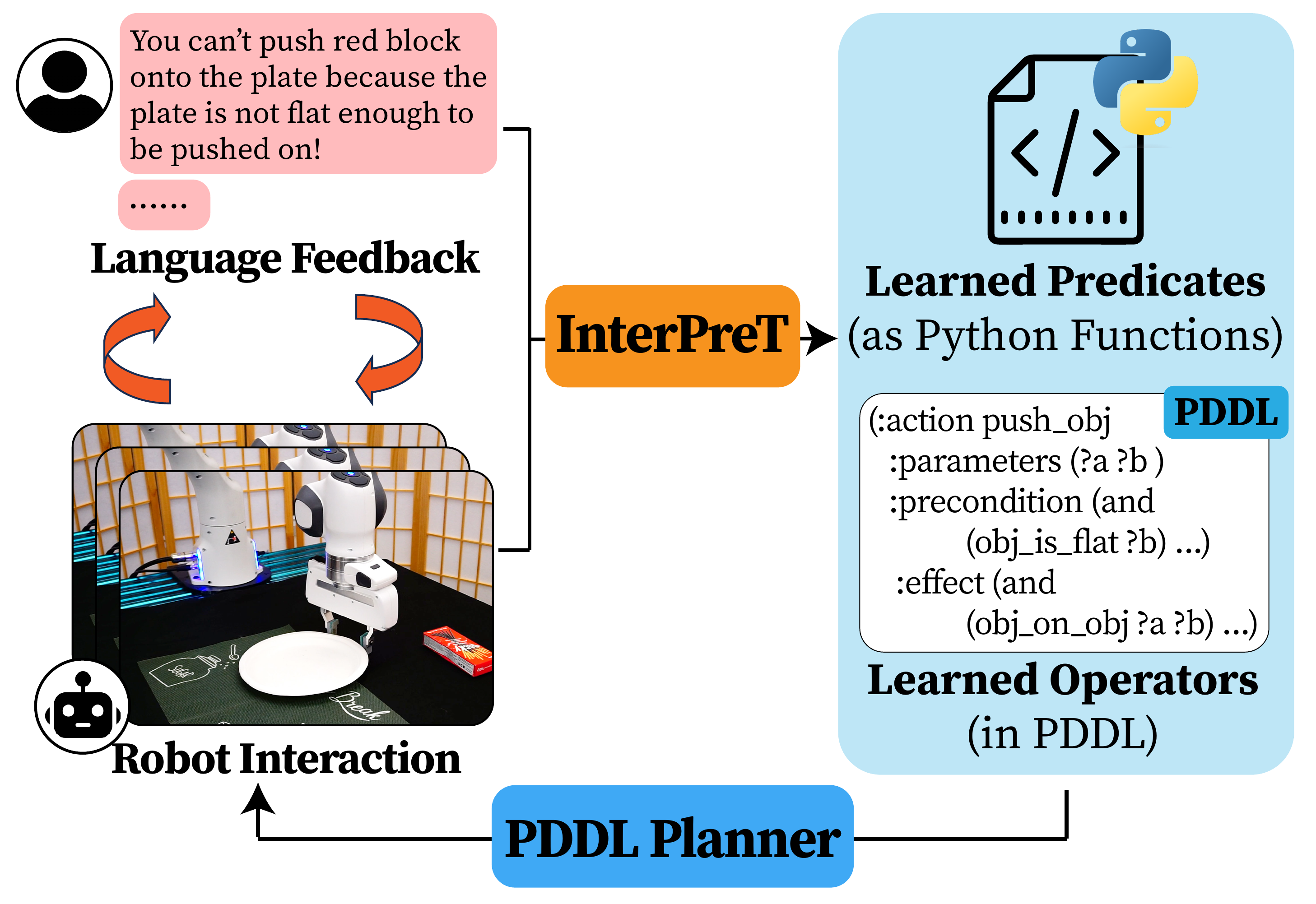}
    \caption{\textbf{\interpret learns predicates as Python functions and operators in PDDL from human language feedback during embodied interaction.} The learned predicates and operators can be leveraged by a PDDL planner for planning for unseen tasks involving more objects and novel goals. }
    \label{fig:teaser}
\end{figure}

We hypothesize that for robots to achieve human proficiency in learning predicates for planning, they must possess an ability similar to infants to learn from the rich human language feedback in an interactive manner.
Recent work has incorporated human language feedback into learning reward functions~\cite{ma2023eureka} and motion policies~\cite{liu2023interactive}. The crux of these methods is to harness the capabilities of pretrained \acp{llm}~\cite{zhao2023survey}, in particular GPT-4~\cite{openai2023gpt4}, in understanding natural language input, performing reasoning~\cite{kojima2022large,yao2022react}, and generating text-based responses (computer programs~\cite{chen2021evaluating}, \etc). Following this line of work, we present \textbf{\interpret} (\textbf{Inter}ative \textbf{Pre}dicate Learning for \textbf{T}ask Planning), the first framework for robots to learn planning-oriented predicates from interactive language feedback, as depicted in \cref{fig:teaser}. \interpret formalizes predicate learning as generating Python functions with GPT-4, which are iteratively refined based on human language feedback. These predicates (as Python functions) can access raw environment states with Python perception APIs and freely compose logic structures and arithmetical computations (\eg, with NumPy) to form complex semantics. With the learned predicates, we can easily learn symbolic operators from the robot's interaction data using a cluster-and-search algorithm~\cite{silver2021learning}. The learned predicates and operators are compiled into the \ac{pddl} format on the fly to be used by a planner. \acp{llm}' capabilities of open-world text processing and symbolic planners' performance guarantees together empower our approach to generalize strongly to arbitrary tasks in the target domains.

Specifically, we consider language feedback for learning two types of planning-oriented predicates, \ie, goal predicates and action precondition predicates~\cite{konidaris2018skills}. These predicates play an essential role in indicating task progress and determining action feasibility, respectively. We design a concise and natural communication protocol to incorporate this feedback:
\begin{itemize}
\item \textbf{Feedback for learning goal predicates}: At the beginning of each task, the human user specifies the goal, \eg, ``put plate on table mat''. Then, it signals when the robot achieves the goal and it explains any unsatisfied conditions if the robot mistakenly declares success.
\item \textbf{Feedback for learning precondition predicates}: The human user verifies the feasibility of the action the robot proposes to execute next. They explain any violated preconditions if the action is infeasible, \eg, ``you can't pick up the plate 
because it is too large for the gripper to grasp'', or otherwise confirm that the action is feasible, \eg, ``you can go ahead and pick up red block''.
\end{itemize}
This protocol allows \interpret to verify and refine the learned predicates from time to time, enabling predicate learning with closed-loop feedback.

In the experiments, we evaluate \interpret's effectiveness in a suite of simulated and real-world robot manipulation domains. These domains are designed such that their dynamics can be modeled using specific predicates and operators, which the robot must uncover. We first have \interpret learn predicates and operators by having the robot interact with a series of simple training tasks while receiving natural language feedback from human users. We then test the learned predicates and operators on harder tasks involving more objects and novel goals. We show with qualitative and quantitative results that: (i) \interpret learns valid predicates and operators that capture essential regularities governing each domain. (ii) The learned predicates and operators allow the robot to solve challenging unseen tasks requiring combinatorial generalization, with a 73\% success rate in simulation, outperforming all baselines by a large margin. (iii) \interpret can effectively handle real-world uncertainty and complexities, operating with considerable performance in real-world robot manipulation tasks.

\section{Related Work}
\label{sec:review}

\subsection{Learning Symbolic Representations for Planning }

Learning symbolic abstractions of complex domains for effective planning is a long-standing pursuit in the planning community~\cite{pasula2007learning,jetchev2013learning,ugur2015bottom,konidaris2018skills,ahmetoglu2022deepsym,silver2023predicate,zhang2023learning,rosen2023synthesizing}. Previous methods have focused on discovering propositional~\cite{konidaris2018skills} or predicate state symbols~\cite{james2022autonomous} from embodied experience. These symbols are usually acquired by composing predefined features~\cite{pasula2007learning,loula2019discovering,curtis2022discovering,silver2023predicate}, or learning statistical~\cite{jetchev2013learning} or neural network~\cite{ahmetoglu2022deepsym} models with clustering~\cite{ugur2015bottom,konidaris2018skills,james2022autonomous} or representation learning techniques~\cite{umili2021learning,asai2019unsupervised,ahmetoglu2022deepsym}. Such learning often relies on unsupervised objectives like minimizing state reconstruction error~\cite{umili2021learning,asai2019unsupervised,ahmetoglu2022deepsym}, prediction error~\cite{jetchev2013learning,umili2021learning,ahmetoglu2022deepsym}, bisimulation distance~\cite{curtis2022discovering} or planning time~\cite{silver2023predicate}. However, these approaches struggle to capture high-level semantic relations~\cite{asai2019unsupervised,ahmetoglu2022deepsym} and often require manual feature engineering~\cite{loula2019discovering,curtis2022discovering,silver2023predicate}.

Supervised learning has also been explored to ground semantic predicates to continuous observations, \eg, images or continuous states~\cite{xu2017scene,mao2019neuro}. While large-scale annotated datasets~\cite{krishna2017visual} are available to learn general-purpose predicate grounding models, fine-tuning with task-specific data is still needed for learned predicates to serve reasoning and planning in specific domains~\cite{zhang2023grounding,kamath2023s,guo2023doremi}. To reduce annotation needs, prior works have employed active learning~\cite{bobu2022learning,lis2023embodied} or novel labeling techniques~\cite{migimatsu2022grounding,mao2022pdsketch}, but a minimum of 500-1000 labels~\cite{bobu2022learning,lis2023embodied} are still required per predicate.

Our work builds on this line of research in learning symbolic abstractions from interaction data and weak supervision. We mitigate limitations of unsupervised methods by learning predicates from natural language feedback. Meanwhile, we are able to learn semantic predicates as Python functions from a few data samples, leveraging the code generation capability and world knowledge of GPT-4.

\subsection{Large Language Models-enabled Planning and Learning }

Large Language Models~\cite{zhao2023survey} have shown remarkable abilities in encoding vast semantic knowledge and demonstrate emergent capabilities in learning, reasoning, and planning with few-shot or even zero-shot prompting~\cite{brown2020language,kojima2022large,huang2022language}. Pretrained \acp{llm} have been applied as planners in text-based environments with natural language instructions and feedback~\cite{yao2022react,huang2022language,li2022pre,shinn2023reflexion,wang2023jarvis}. For \textit{grounded} planning in realistic robotic domains, a common approach is to utilize out-of-the-box perception models to convert raw observations into textual descriptions for \acp{llm} to consume~\cite{song2023llm,wu2023embodied,huang2022inner,du2023vision,wang2024llm3,zhi2024closed}, or provide perception and action APIs for \acp{llm} to generate executable programs~\cite{liang2023code,singh2023progprompt}. However, these perception models struggle to capture complex task-relevant information like semantic object relations without task-specific tuning~\cite{zhang2023grounding,guo2023doremi}. Leveraging GPT-4's power, our work effectively acquires meaningful task-relevant predicates to facilitate grounded planning.

Pretrained \acp{llm} are also leveraged to enhance robot agent intelligence by generating formatted outputs (\eg, code, formal language) and refining them based on language feedback via iterative prompting. They have been used as interfaces between natural language and robotics modalities like formal planning languages~\cite{xie2023translating,liu2023llm+} (\eg, PDDL~\cite{fox2003pddl2}), reward functions~\cite{ma2023eureka,yu2023language} and trajectories~\cite{liu2023interactive}. Specifically, Voyager~\cite{wang2023voyager} uses GPT-4 to construct an automatic curriculum and a skill library to build lifelong learning agents, while Eureka~\cite{ma2023eureka} and OLAF~\cite{liu2023interactive} leverage GPT-4 for learning from language feedback effectively by prompting. Inspired by these works, we learn predicates from language feedback by generating and iteratively refining Python functions with GPT-4.

\begin{figure*}[!ht]
    \centering
    \includegraphics[width=\linewidth]{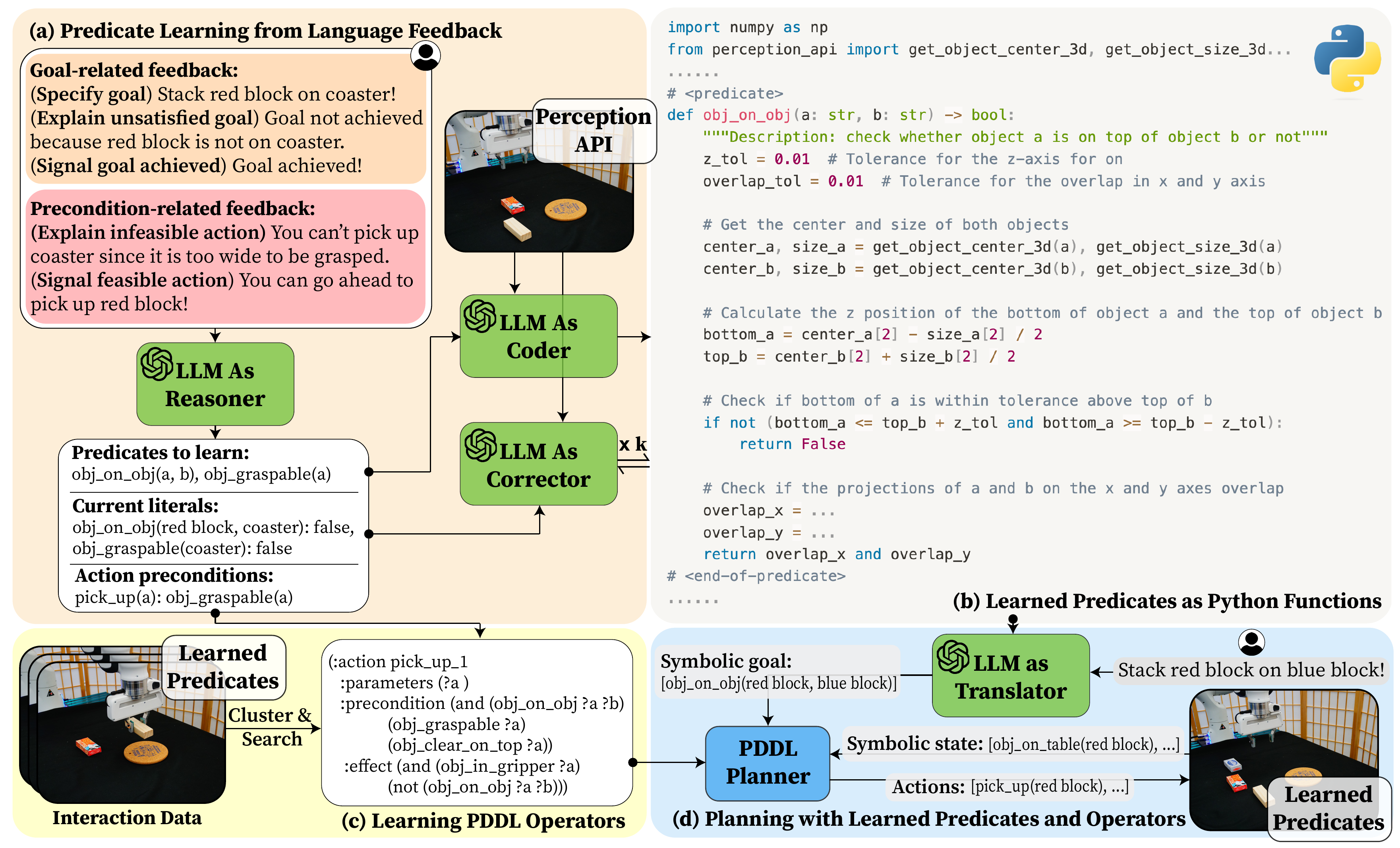}
    \caption{\textbf{The system architecture of \interpret.} (a) We design three GPT-4-enabled modules that operate sequentially to identify planning-oriented predicates and generate the predicate functions based on language feedback. (b) An example predicate function learned. (c) With the learned predicates, we learn PDDL operators with a cluster-then-search algorithm. (d) The learned predicates and operators enable effective task planning, after we translate language goals into symbolic goals with GPT-4.}
    \label{fig:architecture}
\end{figure*}

\section{Preliminaries and Problem Setup}
\label{sec:problem}

We consider robot task planning in a continuous state space $O$ with language goal specifications $G$. Without losing generality, we assume the states are factorized with respect to a set of objects $E$, where such information can be obtained using mainstream perception models like object detectors. The robot is equipped with a library of primitive actions $A$, where each $a \in A$ is parameterized by object variables and can be grounded to certain objects to produce an executable action $\underline{a}$, \eg, \texttt{Pick(cup)}. Then, a task planning problem is to find a sequence of actions $\underline{a}_{1:T}$ to reach a final state $o_g$ that satisfies a language goal $g \in G$ from an initial state $o_0 \in O$.

Following the classical planning formulation~\cite{russell2010artificial,fikes1971strips}, we aim to learn predicates $\Psi$ to abstract the state space $O$ into a symbolic one $S$ for effective and generalizable planning. A predicate $\psi:=<d_{\psi},f_{\psi}> \in \Psi$ defines a function $f_{\psi}$ that captures a symbolic relation among a list of object variables, with its semantic meaning described as $d_{\psi}$. The function $f_{\psi}:O \times E^k \rightarrow \{0,1\}$ takes a continuous state $o \in O$ and a list of $k$ objects $(e_1,e_2,...,e_k) \in E^k$ and outputs a binary value indicating whether the relation holds or not. For example, a predicate \texttt{on(a,b)} can be applied to check whether \texttt{cup} is physically on \texttt{plate}, producing a positive literal \texttt{on(cup, plate)} or a negative literal $\neg \texttt{on(cup, plate)}$. Then the symbolic state $s$ of a continuous state $o$ can be obtained by collecting all positive literals at state $o$ given predicate set $\Psi$ and object set $E$, denoted $s=\texttt{Parse}(o;\Psi,E)$.

With the object-factorized symbolic state space $S$, we model the preconditions and effects of primitive actions with symbolic operators $\Omega$. Each symbolic operator $\omega \in \Omega$ corresponding to a primitive action $a$ is characterized by a precondition set $\text{CON}$ (literals must hold before executing $a$), and adding and deleting effect set $\text{EFF$^+$}$ and $\text{EFF$^-$}$ (literals added and removed from symbolic state $s$ after executing $a$). These symbolic operators are \textit{lifted} by design, enabling the evaluation of preconditions and effects for any executable version $\underline{a}$ obtained by applying the primitive action $a$ to any objects. With the learned predicates $\Psi$, we further learn the symbolic operators $\Omega$ of all primitive actions $A$ to achieve generalizable task planning. The learned predicates and operators can be compiled into a \ac{pddl} domain. By converting a language goal $g \in G$ into a symbolic goal $s_g$~\cite{liu2023llm+,xie2023translating}, such a PDDL domain can enable effective planning using an off-the-shelf classical planner~\cite{helmert2006fast}.

\section{Method}
\label{sec:method}

In this section, we present the \interpret framework that learns predicates and operators from language feedback for planning. The overall architecture is depicted in \cref{fig:architecture}. There are five essential modules that operate together to empower \interpret: (i) \textbf{\textit{Reasoner}}, which analyzes language feedback to identify new predicates and extract task-relevant information (\eg, predicate labels, action preconditions), (ii) \textbf{\textit{Coder}}, which generates Python functions to ground the new predicates, (iii) \textbf{\textit{Corrector}}, which iteratively refines existing predicate functions to align their predictions to the extracted predicate labels, (iv) \textbf{\textit{Operator Learner}}, which learns operators from interaction data based on the learned predicates, and (v) \textbf{\textit{Goal Translator}}, which translates language goal specifications into symbolic goals to enable planning. Below, we elaborate on the core GPT-4-powered modules-\textit{Reasoner}, \textit{Coder} and \textit{Corrector}-that enable predicate learning, and briefly introduce the rest, which are mainly adapted from existing works.

Given language feedback $l_t$ at time step $t$, our objective is to learn new predicates and refine existing predicates $\Psi_{t-1}$, producing an updated set of predicates $\Psi_t$. For simplicity of notation, we denote the textual descriptions of predicates as $\{d_{\psi}\}$ and the corresponding predicate functions as $\{f_{\psi}\}$ for any predicate set $\Psi$. We decompose the predicate learning process at time step $t$ into three sequential sub-steps (see \cref{fig:architecture}(a)): (i) \textit{Reasoner} identifies new predicates with descriptions $\{d_{\psi_{new}}\}$ and extracts current state literals that provide predicate labels $\{y\}$, (ii) \textit{Coder} generates new predicate functions $\{f_{\psi_{new}}\}$, and (iii) \textit{Corrector} refines existing predicate functions to fix execution errors and match their predictions to $\{y\}$. Formally, we summarize this process in \cref{eq:llm}:
\begin{equation}
    \begin{aligned}
        \text{Reasoner~:~} & \{d_{\psi_{new}}\}, \{y\} = f_{Reason}(l_t,\{d_{\psi_{t-1}}\}), \\
        & \{d_{\psi_{t}}\} = \{d_{\psi_{new}}\} \cup \{d_{\psi_{t-1}}\}\\
        \text{Coder~:~} & \{f_{\psi_{new}}\} = f_{Code}(\{d_{\psi_{new}}\},\Psi_{t-1}), \\
        & \{f_{\hat{\psi}_t}\} = \{f_{\psi_{new}}\} \cup \{f_{\psi_{t-1}}\}, \\
        \text{Corrector~:~} & \{f_{\psi_t}\} = f_{Correct}(o_t, \{y\}, \{f_{\hat{\psi}_t}\}), \\
        & \Psi_t = \{<d_{\psi_t},f_{\psi_t}>\},
    \end{aligned}
    \label{eq:llm}
\end{equation}
where $f_{Reason}$, $f_{Code}$, $f_{Correct}$ are parameterized by GPT-4 with varying prompt templates, and the initial predicate set is empty, \ie, $\Psi_0 = \emptyset$. Note that we omit some of the output terms irrelevant to predicate learning for clarity. We detail the modules below and provide the complete prompt templates in the supplementary material.

\subsection{Reasoner}
\label{subsec:reasoner}

The \textit{Reasoner} module is designed to identify essential predicates and extract task-relevant information from goal-related or precondition-related language feedback; see the top left corner of \cref{fig:architecture}(a) for examples of these feedback types. We tailor \textit{Reasoner} to each language feedback type using different prompt templates, as detailed below. We highlight \hllb{the language feedback in blue}, and \hllo{the GPT-4 output in orange}. We employ \ac{cot} prompting~\cite{wei2022chain} for \textit{Reasoner} to provide the complete reasoning trace and in-context learning~\cite{kojima2022large} to enable \textit{Reasoner} to learn from a single example. These techniques are applied to all LLM promptings to facilitate robust response generation. 

\subsubsection{Goal-related feedback}
\begin{itemize}
    \item \textbf{(Specify task goal)} Given a natural language goal specification, \textit{Reasoner} identifies new goal predicates and converts the language goal into symbolic form using existing and new predicates:

    \framedtext{
    \textrm{\small 
    \textbf{Context:} ...\{example\} \{objects\} \{existing\_predicates\}... \\
    \hllb{\textbf{Goal:} Stack red block on coaster.} \\
    \hllo{\textbf{Reasoning:} The goal can be captured by a symbolic literal obj\_on\_obj(red block, coaster). As predicate obj\_on\_obj(a, b) is unknown, we need to learn it. \\
    \textbf{Predicates to learn:} \{``obj\_on\_obj(a, b)'': ``check whether object a is on object b''\} \\
    \textbf{Symbolic goal}: \{``obj\_on\_obj(red block, coaster)'': true\}}}
    }
    The identified predicates to learn $\{d_{new}\}$ are to be supplied to \textit{Coder} for predicate function generation, and the converted symbolic goal $s_g$ is recorded for robot planning and exploration.
    
    \item \textbf{(Explain unsatisfied goal)} Given explanations for unsatisfied goal literals, \textit{Reasoner} extracts the current symbolic literals from the language feedback, which provide predicate labels $\{y\}$. These predicate labels are crucial for \textit{Corrector} to correct the generated goal predicates.

    \framedtext{
    \textrm{\small 
    \textbf{Context:} ...\{example\} \{objects\} \{existing\_predicates\}... \\
    \hllb{\textbf{Human explanation:} You haven't reached the goal because red block is not on coaster.} \\
    \hllo{\textbf{Current symbolic state:} \{``obj\_on\_obj(red block, coaster)'': false\}}}
    }

    \item \textbf{(Signal goal achieved)} When receiving goal-achieved signals, we simply use the extracted symbolic goal as the current literals, which provide positive labels $\{y\}$ to supervise goal predicates. 
\end{itemize}

\subsubsection{Precondition-related feedback} 

\begin{itemize}
    \item \textbf{(Explain infeasible action)} Given an explanation of an infeasible action, \textit{Reasoner} identifies new precondition predicates to learn $\{d_{new}\}$, reasons about the current symbolic state $\{y\}$, and summarizes the reflected action preconditions $\text{CON}_{new}$, as shown in the example below. The summarized action preconditions only take arguments that exist in those of the action, \eg, for action \texttt{pick\_up(a)}, we only summarize the preconditions that take no argument or object \texttt{a} as an argument. We aggregate $\text{CON}_{new}$ into the precondition set $\text{CON}_{a}$ of the corresponding action $a$, which is maintained for operator learning (described in \cref{subsec:other}). Additionally, $\{d_{new}\}$ is fed into \textit{Coder} to generate precondition predicate functions, and the predicate labels $\{y\}$ are provided to \textit{Corrector} for correction.

    \framedtext{
    \textrm{\small 
    \textbf{Context:} ...\{example\} \{objects\} \{existing\_predicates\}... \\
    \textbf{Infeasible action:} pick\_up(coaster) \\
    \hllb{\textbf{Human explanation:} You can't pick up coaster it is too large to be grasped.} \\
    \hllo{\textbf{Reasoning:} ... The precondition of pick\_up(coaster) is that it is small enough to be grasped by the gripper... We learn predicate obj\_graspable(a) to check whether object a can be grasped by the gripper...\\
    \textbf{Predicates to learn:} \{``obj\_graspable(a)'': ``check whether object a is small enough to be grasped by the gripper...''\}\\
    \textbf{New action preconditions:} \{``action'': "pick\_up(a)'', ``new preconditions'': \{``obj\_graspable(a)'': true\}\} \\
    \textbf{Current symbolic state:} \{``obj\_graspable(coaster)'': false\}
    }
    }
    }

     \item \textbf{(Signal feasible action)} When an action is signaled as feasible, we confirm that all preconditions in $\text{CON}_{a}$ are satisfied. The labels ${y}$ for these precondition predicates are obtained and provided to \textit{Corrector} to correct precondition predicate functions.
\end{itemize}

\subsection{Coder}
\label{subsec:coder}

Once \textit{Reasoner} identifies a set of new predicates with text descriptions $\{d_{new}\}$, the next step is to construct the corresponding predicate functions $\{f_{new}\}$ to truly learn them. Inspired by recent successes in using pretrained LLMs to generate computer programs for robotic tasks~\cite{liang2023code,singh2023progprompt,wang2023voyager,ma2023eureka}, we design \textit{Coder} to generate these predicate functions as Python code based solely on $\{d_{new}\}$, leveraging the power of GPT-4. We assume the availability of a library of perception API functions that provide access to continuous states, such as the bounding boxes and categories of detected objects. The predicate functions can then be constructed by composing these API functions with classical logic structures and arithmetical computations (\eg, using NumPy), exploiting the flexibility of Python programming. Representing predicates as Python functions offers several advantages: (i) They are semantically rich and interpretable compared to neural networks~\cite{ahmetoglu2022deepsym,umili2021learning}, and have better representation power and more versatile syntax than logical programs~\cite{silver2023predicate,curtis2022discovering}. (ii) They enable one-shot generation purely from the text description without labeled data, leveraging the extensive commonsense priors in GPT-4.

To facilitate the construction of predicate functions, we provide \textit{Coder} with the following primitives: (i) perception API functions for accessing environment states, (ii) the NumPy library for arithmetic computations, and (iii) if-else and loop statements for controlling the logic structure. We also allow \textit{Coder} to create additional utility functions that can be reused to define different predicate functions. This divide-and-conquer strategy helps mitigate the complexity of building predicate functions from scratch. In practice, we prompt GPT-4 with a code snippet demonstrating the usage of primitives by a few examples of utility functions and one example predicate function. Detailed comments are included in these examples to enable \ac{cot} prompting. Due to space limitations, we show a partial prompt with an example utility function \texttt{\small get\_object\_x\_range} below, and an example predicate function in \cref{fig:architecture}(b). The complete prompt for \textit{Coder} can be found in the supplementary material.

\begin{codeframe}
\footnotesize
\begin{lstlisting}[style=PythonStyle]
import numpy as np
from perception_api import get_object_center_3d, get_object_size_3d...

# <utility>
def get_object_x_range(a) -> np.ndarray:
    """
    Get the range of object a along x axis
    :param a: string, name of detected object
    """
    center = get_object_center_3d(a)
    extent = get_object_size_3d(a)
    return np.array([center[0] - extent[0] / 2, center[0] + extent[0] / 2])
# <end-of-utility>

......
\end{lstlisting}
\end{codeframe}

\subsection{Corrector}
\label{subsec:corrector}

The predicate functions generated by \textit{Coder} are pruned to two types of errors: (i) \textbf{execution errors} due to invalid operations or syntax errors in the function, and (ii) \textbf{alignment errors} that arise when GPT-4's understanding of predicates misaligns with that of the human user. Inspired by \cite{wang2023voyager,ma2023eureka}, we introduce \textit{Corrector}, which adopts an iterative prompting mechanism to correct the functions and rectify these errors. 

When an execution error occurs while calling a predicate function, \textit{Corrector} takes the execution trace from the code interpreter and the erroneous code as input. It then performs zero-shot reasoning on the error and generates the corrected code as output. This process is repeated until the error is resolved or the maximum number of iterations is reached. Similarly, to address alignment errors, \textit{Corrector} is provided with the predicate labels $\{y\}$ and the current predicate functions. It then iteratively refines the functions to align their outputs with the given labels. An example prompt illustrating this process is shown below:

\framedtext{
\textrm{\small 
\textbf{Context:} ...\{instructions\} \texttt{<code>} \{observation\}... \\
\hllb{\textbf{Alignment error:} obj\_on\_obj(red block, blue block) should be true, but your prediction is false.} \\
\hllo{\textbf{Reasoning:} This is because we assume object a should be within object b along the x-axis when we define the predicate function obj\_on\_obj(a, b), which is wrong; in fact, the relation is established when object a is overlapped with b, so we need to correct the function accordingly.   \\
\textbf{Corrected code:} \texttt{<corrected\_code>}
}
}}

\begin{figure*}[th!]
    \centering
    \includegraphics[width=\linewidth]{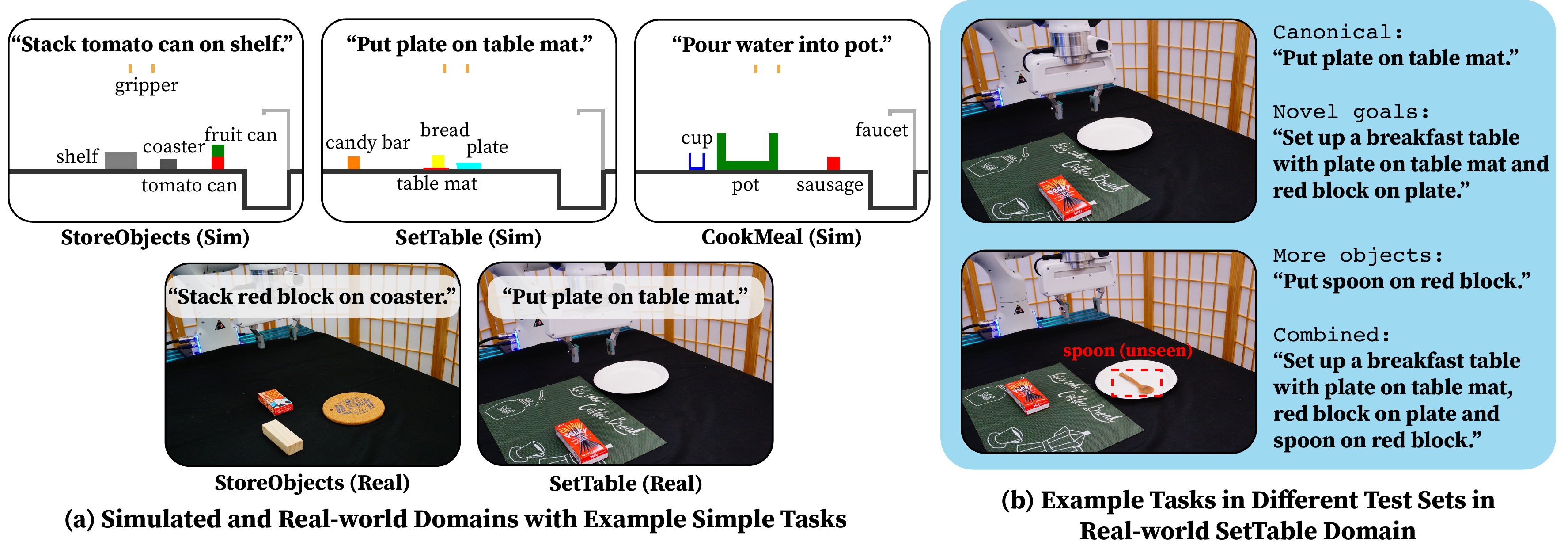}
    \caption{\textbf{Simulated and real-world domains used in the experiments.} We show example training tasks of all five domains in (a) and demonstrate the design of the 4 test sets in the real-world SetTable domain in (b). In \texttt{More objects} and \texttt{Combined}, an unseen object ``spoon'' introduces additional generalization challenges.}
    \label{fig:domains}
\end{figure*}

\subsection{Other Components}
\label{subsec:other}

Given the learned predicates, we implement a variant of the cluster-then-search algorithm~\cite{silver2021learning} to learn operators. This algorithm effectively learns symbolic operators that best capture the action effects and preserve a minimal set of necessary preconditions from a small number of successful and failed interactions. We also incorporate the action preconditions summarized by \textit{Reasoner} into the learned operators. To ensure learning from language feedback with no delays, we run the operator learning algorithm at each interaction step, maintaining an operator set compatible with the up-to-date predicates and interaction experience. The implementation details are included in the supplementary material.

During the training phase, \interpret learns predicates and operators as the robot interacts with the environment to perform a series of training tasks (detailed in \cref{subsubsec: eval}). We employ a strategy where the robot plans with a classical planner~\cite{helmert2006fast} based on the learned predicates and operators 50\% of the time, and randomly takes a symbolically feasible action according to the recorded action preconditions otherwise. Empirically, this approach enables a balance between exploration and exploitation.

At test time, we introduce an \ac{llm}-based goal translator to convert language goals into symbolic form, following previous works~\cite{xie2023translating,liu2023llm+}. We refer the reader to the original papers for a detailed explanation of how the method works. In practice, we find that the GPT-4-based goal translator performs robustly when provided with a few examples. 

\section{Experiments}
\label{sec:exp}

We conduct experiments to answer the following questions: (i) Can \interpret learn meaningful task-relevant predicates and operators from language feedback? (ii) How well do the learned predicates and operators (\ie, \ac{pddl} domains) generalize to tasks that involve more objects and novel goals? (iii) Can \interpret handle perception and execution uncertainties in the real world?

\subsection{Experimental Setup}
We quantitatively and qualitatively evaluate \interpret on a suite of robot manipulation domains in a simulated 2D kitchen environment~\cite{wang2018active} and a real-world environment. The domain design, baseline methods, and evaluation protocol are described below.

\subsubsection{Domain design}
We design three simulated domains based on the Kitchen2D environment~\cite{wang2018active} and two real-world domains that represent the counterpart of the simulation. Each domain is associated with a set of simple and complex tasks, and designed with a ground-truth PDDL domain file specifying the essential symbolic constraints and regularities. The five domains are each demonstrated with an example simple task in \cref{fig:domains}(a). We briefly introduce the domains below and include further details in the supplementary material. 

\begin{itemize}
    \item \textbf{StoreObjects (Sim and Real)}: This domain involves storing objects on a large receptacle by picking, placing, and stacking actions. It features predicates and corresponding constraints similar to those in the BlockWorld domain~\cite{gupta1991complexity}, such as \texttt{on(a,b)}, and \texttt{on\_table(a)}.
    \item \textbf{SetTable (Sim and Real)}: This domain involves rearranging objects to set up a breakfast table. Compared to StoreObjects, it additionally introduces a push action to move large objects (\eg, plates) that cannot be grasped. It features precondition predicates such as \texttt{is\_graspable(a)} and \texttt{is\_flat(a)}.
    \item \textbf{CookMeal (Sim only)}: This domain involves putting ingredients into a pot and filling the pot/cups with water. It requires understanding the task semantics of actions, featuring predicates such as \texttt{is\_container(a)}, \texttt{in(a,b)} and \texttt{has\_water(a)}. It also imposes constraints such as the only way to fill a large container (\eg, a pot) with water is by using a cup.
\end{itemize}

\subsubsection{Baselines}
As there are no prior methods that learn predicates from human language feedback for planning, we compare \interpret with state-of-the-art \ac{llm}-based planners. (i) \textbf{Inner Monologue (IM)}\cite{huang2022inner} generates action plans based on textualized environment states using an LLM. (ii) \textbf{Code-as-Policies (CaP)}\cite{liang2023code} employs an LLM to generate policy code that invokes perception and action APIs. We also implement variants of IM that incorporate predicates and operators learned with \interpret. For a fair comparison, all baselines access the environment state through perception APIs and learn from in-context examples.
\begin{itemize}
    \item \textbf{IM\,+\,Object}~\cite{huang2022inner}: A naive IM variant that utilizes the textualized output of perception APIs, \eg, detected objects with positions and categories, as the environment state.
    \item \textbf{IM\,+\,Object\,+\,Scene}~\cite{huang2022inner}: An IM variant that uses environment states augmented by scene descriptions, obtained using predicates learned by \interpret.
    \item \textbf{IM\,+\,Object\,+\,Scene\,+\,Precond}~\cite{huang2022inner}: An IM variant that leverages the operators learned with \interpret to check the precondition of actions proposed by IM. Infeasible actions are prompted back to the LLM for replanning.
    \item
    \textbf{CaP}~\cite{liang2023code}: A strong CaP baseline that performs precondition checks using ``assertion'' or if-else statements (akin to ProgPrompt\cite{singh2023progprompt}) and hierarchically composes policies for long-horizon planning. We have it generate predicate functions for precondition checks.
\end{itemize}

\subsubsection{Evaluation protocol}
\label{subsubsec: eval}
We adopt a train-then-test evaluation workflow for all domains. For each domain, the robot first learns from a series of 10 simple training tasks accompanied by language feedback. For testing, we design four sets of tasks (10 tasks per set) that pose different levels of challenge to the generalizability of the methods. We present example tasks in different test sets of the real-world SetTable domain in \cref{fig:domains}(b). 
\begin{itemize}
    \item \texttt{Canonical}: Simple tasks with objects and goals seen in training but with different initial configurations.
    \item \texttt{More objects}: Simple tasks with seen goals but involve additional unseen objects.
    \item \texttt{Novel goals}: Complex tasks with seen objects but novel goals that compose goals seen in training tasks.
    \item \texttt{Combined}: Complex tasks with unseen objects and goals, combining the last two setups.
\end{itemize}
We evaluate the performance of all methods using the success rate on the 10 tasks of each test set. In simulation, we conduct systematic evaluations by running the whole training-testing pipeline 3 times with varied seeds. We directly terminate the episode upon action failure for all methods.

\begin{table}[th!]
\centering
{\scriptsize
\begin{tabular}{l|cc@{}}
\toprule
\textbf{Domain}       & \textbf{Goal Predicates}               & \textbf{Precondition Predicates} \\ \midrule
StoreObjects  & \makecell{\texttt{obj\_on\_obj(a, b)}, \\ \texttt{obj\_on\_table(a)}} & \makecell{\texttt{obj\_graspable(a)}, \\ \texttt{obj\_clear(a)},\\ \texttt{gripper\_empty()}} \\ \hline
SetTable      & \makecell{\texttt{obj\_on\_obj(a, b)}, \\ \texttt{obj\_on\_table(a)}} & \makecell{\texttt{obj\_graspable(a)}, \\ \texttt{obj\_clear(a)}, \\ \texttt{gripper\_empty()}, \\ \texttt{obj\_is\_plate(a)}, \\ \texttt{obj\_thin\_enough(a)}} \\ \midrule
CookMeal      & \makecell{\texttt{obj\_inside\_obj(a, b)}, \\ \texttt{obj\_on\_table(a)}, \\ \texttt{obj\_filled\_with\_water(a)}} & \makecell{\texttt{obj\_graspable(a)}, \\ \texttt{obj\_clear(a)}, \\ \texttt{gripper\_empty()}, \\ \texttt{obj\_is\_plate(a)}, \\ \texttt{obj\_thin\_enough(a)}, \\ \texttt{obj\_large\_enough(a)}, \\ \texttt{obj\_is\_food(a)}, \\ \texttt{obj\_is\_container(a)}} \\ \bottomrule
\end{tabular}
\caption{\textbf{Learned goal and precondition predicates in simulated domains.} We report the union of the three runs. While we learn both the positive predicate and its negated counterpart, we only show the positive ones here for clarity. We adjust some of the predicate names to unify them across domains and runs for better readability.  }
\label{table:predicates}
}
\end{table}

\begin{figure}[th!]
    \centering
    \includegraphics[width=\linewidth]{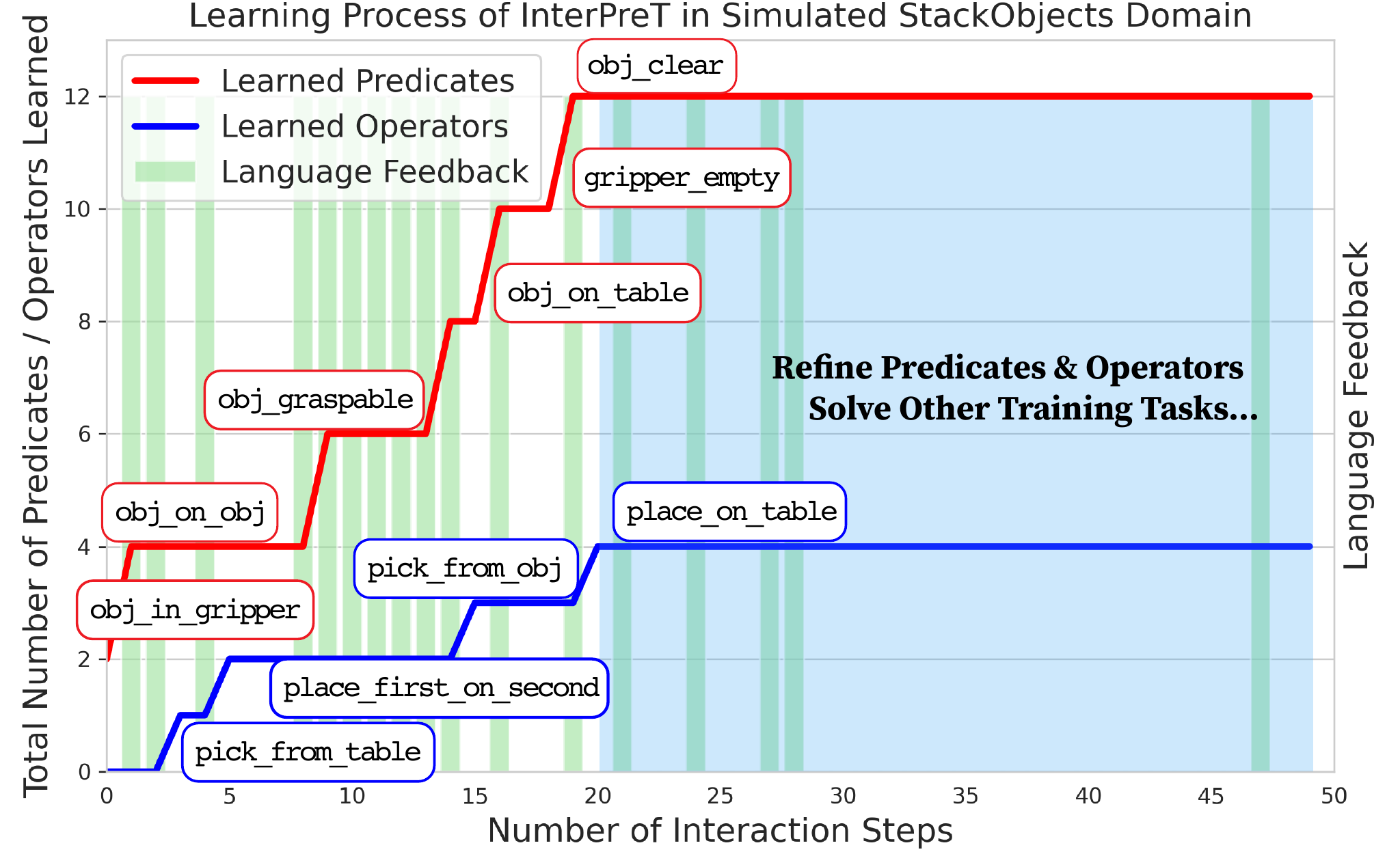}
    \caption{\textbf{Visualization of one training run in simulated StoreObjects domain.} The total number of learned predicates increases by 2 for each labeled predicate as we also learn its negation. We provide the predicate function of \texttt{obj\_in\_gripper} as an in-context example, known at Step 0. We empirically label the learned operators with semantic names based on their interpreted meanings. }
    \label{fig:process}
\end{figure}

\begin{table*}[ht]
\centering
{\small
\begin{tabular}{@{}l l | c x x | c | y}
\toprule
Domain & Test Set & IM\,+\,Object~\cite{huang2022inner} & IM\,+\,Object & IM\,+\,Object & CaP~\cite{liang2023code} & \textbf{\interpret (Ours)} \\
& & & \,+\,Scene~\cite{huang2022inner} & \,+\,Scene\,+\,Precond~\cite{huang2022inner} & & \\
\midrule
StoreObjects & \texttt{Canonical}         & $0.60 \pm 0.00$ & $0.90 \pm 0.00$ & $\boldsymbol{1.00 \pm 0.00}$ & $\boldsymbol{1.00 \pm 0.00}$ & $0.93 \pm 0.12$ \\
      & \texttt{More objects}      & $0.30 \pm 0.00$ & $0.83 \pm 0.06$ & $\boldsymbol{1.00 \pm 0.00}$ & $0.83 \pm 0.15$ & $0.90 \pm 0.17$ \\
      & \texttt{Novel goals}      & $0.00 \pm 0.00$ & $0.87 \pm 0.15$ & $0.97 \pm 0.06$ & $0.53 \pm 0.21$ & $\boldsymbol{1.00 \pm 0.00}$ \\
      & \texttt{Combined}          & $0.00 \pm 0.00$ & $0.77 \pm 0.06$ & $0.87 \pm 0.15$ & $0.03 \pm 0.06$ & $\boldsymbol{1.00 \pm 0.00}$ \\
\midrule
SetTable & \texttt{Canonical}         & $0.80 \pm 0.10$ & $0.80 \pm 0.10$ & $\boldsymbol{1.00 \pm 0.00}$ & $0.87 \pm 0.06$ & $\boldsymbol{1.00 \pm 0.00}$ \\
      & \texttt{More objects}      & $0.73 \pm 0.06$ & $0.83 \pm 0.12$ & $\boldsymbol{1.00 \pm 0.00}$ & $0.73 \pm 0.15$ & $\boldsymbol{1.00 \pm 0.00}$ \\
      & \texttt{Novel goals}       & $0.00 \pm 0.00$ & $0.10 \pm 0.10$ & $0.53 \pm 0.33$ & $\boldsymbol{0.77 \pm 0.25}$ & $0.53 \pm 0.41$ \\
      & \texttt{Combined}          & $0.00 \pm 0.00$ & $0.03 \pm 0.05$ & $0.20 \pm 0.16$ & $0.33 \pm 0.15$ & $\boldsymbol{0.37 \pm 0.45}$ \\
\midrule
CookMeal & \texttt{Canonical}         & $0.90 \pm 0.00$ & $\boldsymbol{1.00 \pm 0.00}$ & $\boldsymbol{1.00 \pm 0.00}$ & $0.97 \pm 0.06$ &$0.97 \pm 0.06$ \\
      & \texttt{More objects}      & $1.00 \pm 0.00$ & $\boldsymbol{1.00 \pm 0.00}$ & $\boldsymbol{1.00 \pm 0.00}$ & $0.93 \pm 0.06$ & $\boldsymbol{1.00 \pm 0.00}$ \\
      & \texttt{Novel goals}       & $0.97 \pm 0.06$ & $0.93 \pm 0.06$ & $\boldsymbol{1.00 \pm 0.00}$ & $\boldsymbol{1.00 \pm 0.00}$ & $\boldsymbol{1.00 \pm 0.00}$ \\
      & \texttt{Combined}          & $0.00 \pm 0.00$ & $0.23 \pm 0.15$ & $\boldsymbol{0.97 \pm 0.06}$ & $0.77 \pm 0.12$ & $0.83 \pm 0.12$ \\
\midrule
\multicolumn{2}{@{}l|}{Average success rate over \texttt{Combined}} & $0.00$ & $0.34$ & $0.68$ & $0.38$ & $\boldsymbol{0.73}$ \\
\bottomrule
\end{tabular}
}
\caption{\textbf{Systematic evaluations of the methods on all test sets in simulated domains.} We highlight our method in deep gray and baselines that benefit from our learned predicates and/or operators in light grey. \interpret achieves a 73\% success rate in the most challenging \texttt{Combined} test set, outperforming all baselines by a large margin.}
\label{table:results}
\end{table*}

\subsection{Experimental results}

\subsubsection{Qualitative analysis} 
We answer Question (i) by qualitatively analyzing the predicates and operators learned by \interpret in the simulated domains. The full details of the learned predicate functions and operators are included in the supplementary material. \cref{table:predicates} shows \interpret can effectively learn language-grounded and semantically meaningful goal and precondition predicates in all three domains. We report the union of learned predicates over three runs; we observe that the learned predicates are generally consistent across the runs. Specifically, \interpret successfully learns goal predicates that acquire the desired task outcomes, such as ``fruit can on shelf'' and ``plate on table'' in StoreObjects and SetTable domains and ``sausage in pot'' and ``cup is filled'' in CookMeal domain. The learned precondition predicates acutely capture the essential task constraints, such as ``fruit can can only be picked up when there is nothing on its top'', and ``water can only be poured into a container''. These well-learned predicates necessarily build the foundations for learning good operators.

We conduct a case study on one training run in the StoreObjects domain. \cref{fig:process} visualizes the process of learning new predicates and operators (represented as red and blue lines, respectively) while provided with intermittent language feedback (indicated by light green bars). Note that feedback less important for predicate learning, \ie, signaling task success or feasible action, are omitted in the figure for clarity.
We observe that \interpret is able to explore effectively and acquire all predicates in 20 steps of interaction.
Based on the predicates learned, \interpret sequentially learns four operators that exhibit clear semantic meaning. Notably, it recovers two operators \texttt{pick\_from\_table} and \texttt{place\_on\_table} for the same primitive action \texttt{place\_up} that is executed in different contexts.
As the robot blindly explores the domain with inadequate knowledge and continuously proposes infeasible actions, dense language feedback is provided to explain precondition violations in Steps 8-20.
Once \interpret captures all action preconditions, the robot can freely navigate the environment without human intervention.
\cref{fig:process} shows that all predicates and operators are properly initialized at Step 20 and are further corrected and refined in subsequent interactions.

\subsubsection{Evaluating planning and generalization}
We systematically evaluate the planning performance of all methods on the four test sets for each simulated domain. \cref{table:results} presents the full results, demonstrating the strong generalizability of \interpret when planning with a classical planner~\cite{helmert2006fast}. Note that several baselines utilize predicates and operators learned with \interpret; their results are shown in light gray, while \interpret's results are in dark gray. \interpret achieves success rates over 90\% on most test sets. On the challenging \texttt{Combined} test set, which requires strong compositional generalizability, it attains an average success rate of 73\%, substantially outperforming IM variants (IM\,+\,Object, IM\,+\,Object\,+\,Scene, and IM\,+\,Object\,+\,Scene\,+\,Precond) by 73\%, 39\%, and 5\%, respectively, and the CaP baseline by 35\%.

We find the predicates and operators learned with \interpret enable significantly improved generalization in planning, by providing meaningful relational abstractions and explicit transition modeling. The naive IM variant (IM\,+\,Object) struggles to generalize with only textualized state descriptions, solving 0\% of \texttt{Combined} tasks. However, augmenting states with predicates learned by \interpret (IM\,+\,Object\,+\,Scene) boosts the success rate on \texttt{Combined} tasks from 0\% to 34\%. Further ensuring action validity using learned operators (IM\,+\,Object\,+\,Scene\,+\,Precond) rivals \interpret at 68\% average success. This hybrid approach benefits from combining world knowledge in the LLM with validity guarantees from operators. However, we observe that it sometimes fails to reach the goal within the maximum number of steps due to frequent replanning. In contrast, \interpret perform explicit PDDL planning with learned predicates and operators, and thus can generate optimal long-sequence plans with guarantee.

We demonstrate the importance of learning from language feedback by comparing \interpret with CaP, a baseline that generates predicate functions for precondition checks and composes policies for long-horizon planning, but without leveraging language feedback. Although CaP can generate policy code with correct logic based on in-context examples, it occasionally fails to generate accurate predicate functions due to the lack of language supervision. This limitation becomes evident in CaP's poor performance on \texttt{Combined} tasks in the StoreObjects and SetTable domains, which require precise predicate understanding for successful long-horizon planning. The superior performance of \interpret in these challenging scenarios highlights the significant benefits of incorporating natural language supervision compared to CaP.

Furthermore, we explore the transferability of learned predicates to facilitate learning and planning in new domains. We investigate the unsatisfactory performance of \interpret in the SetTable domain, and find that simultaneously learning predicates related to pick-and-place and push actions poses a significant challenge. To address this issue, we bootstrap the learning process with predicates acquired from simpler domains. \cref{table:bootstrap} demonstrates that initializing \interpret with predicates learned in the StoreObjects domain leads to near-perfect learning in the SetTable domain, achieving 100\% success across all test sets. This finding highlights the potential for reusing previously learned predicates to enhance learning efficiency and planning performance in complex domains.

\begin{table}[t!]
\small
\centering
\begin{tabular}{lcc}
\toprule
& From scratch & Bootstrapped \\ \midrule
\texttt{Canonical} & $\boldsymbol{1.00 \pm 0.00}$ & $\boldsymbol{1.00 \pm 0.00}$ \\
\texttt{More objects} & $\boldsymbol{1.00 \pm 0.00}$ & $\boldsymbol{1.00 \pm 0.00}$ \\
\texttt{Novel goals} & $0.53 \pm 0.41$ & $\boldsymbol{1.00 \pm 0.00}$ \\
\texttt{Combined} & $0.37 \pm 0.45$ & $\boldsymbol{1.00 \pm 0.00}$ \\
\bottomrule
\end{tabular}
\caption{\textbf{Bootstrapping predicate learning from previously learned predicates in similar domains.} Reusing predicates learned in StoreObjects leads to near-perfect predicate learning in SetTable. The transfer of predicates is natural as all predicate functions utilize the same Perception API functions.}
\label{table:bootstrap}
\end{table}

\subsubsection{Real-robot results}
\begin{figure}[t!]
    \centering
    \includegraphics[width=\linewidth]{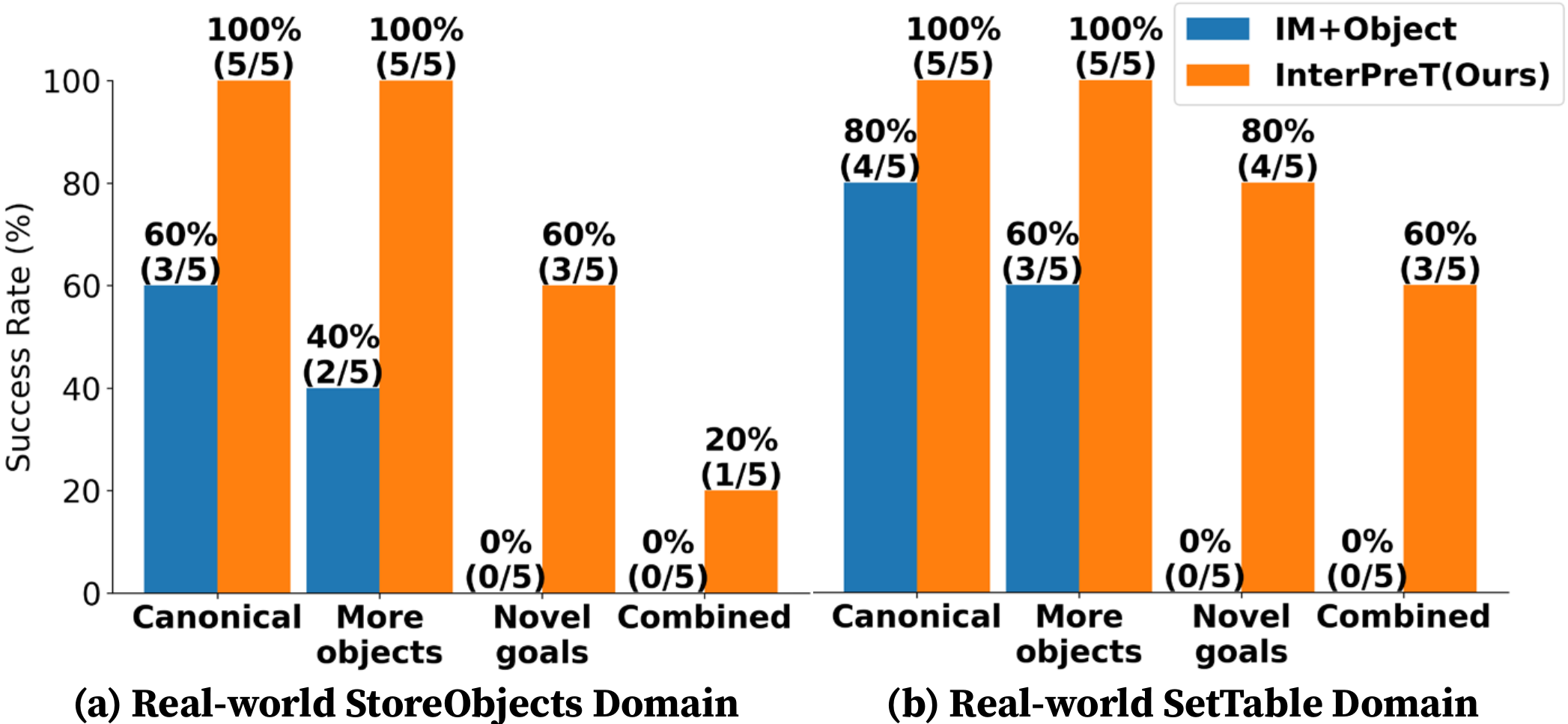}
    \caption{\textbf{Real-robot evaluations in real-world StoreObjects and SetTable domains.} We train \interpret once on 10 tasks and test on 5 tasks per test set. Note that the predicate learning in SetTable is bootstrapped from predicates learned in StoreObjects. }
    \label{fig:real_result}
\end{figure}

We evaluate \interpret in the real-world StoreObjects and SetTable domains, compared to the vanilla IM\,+\,Object baseline~\cite{huang2022inner}. We train \interpret on 10 training tasks while a human user provides language feedback with a keyboard. We then test all methods on 5 test tasks per test set, with the success rates shown in \cref{fig:real_result}. In the SetTable domain, we directly bootstrap the learning with predicates learned from StoreObjects, as the simulated results have already demonstrated the difficulty of learning from scratch in SetTable. The results indicate that \interpret can effectively capture symbolic constraints and regularities in real-world settings where perception and execution uncertainties present. In contrast, the baseline struggles to generalize to novel task goals, highlighting the importance of the learned predicates and operators.
We observe a severe performance drop for \interpret in the StoreObjects domain under the \texttt{Combined} and \texttt{Novel goals} settings. We find that this is attributed to the increased occurrence of primitive action execution failures as the task horizon extends. Despite this, \interpret still outperforms the baseline by a significant margin, achieving success rates of 60\% and 20\% in the \texttt{Novel goals} and \texttt{Combined} settings, respectively.

\subsection{Additional Analysis and Discussions}

\begin{table}[ht]
\small
\centering
\begin{tabular}{ll|ccc}
\toprule
\multicolumn{2}{c|}{\multirow{2}{*}{Stage}} & \multicolumn{3}{c}{Run time / Iteration} \\ \cmidrule{3-5}
& & Median & Min & Max \\ \midrule
\multirow{2}{*}{Training} & Learn predicates & $2.94$ s & $1.05$ s & $23.32$ s \\
& Learn operators & $32$ ms & $1$ ms & $97$ ms \\ \midrule
\multirow{2}{*}{Testing} & Translate goal & $1.60$ s & $0.91$ s & $5.53$ s \\
& Plan with PDDL & $99$ ms & $75$ ms & $130$ ms\\
\bottomrule
\end{tabular}
\caption{\textbf{Run time breakdown of \interpret at different training and testing stages.} We show the median, minimum, and maximum values as the statistics are not normally distributed.}
\label{table:runtime}
\end{table}

\begin{table}[ht]
\centering
{\small
\begin{tabular}{l | c c c}
\toprule
Domain & \textbf{\#LLM Calls} & \textbf{\#Transitions} & \textbf{\#Feedback} \\
\midrule
StoreObjects & $22/31/23$ & $54/75/90$ & $17/26/18$ \\
SetTable & $38/38/62$ & $41/39/67$ & $31/30/55$ \\
CookMeal & $32/29/46$ & $62/34/48$ & $23/22/38$ \\
\bottomrule
\end{tabular}
}
\caption{Number of LLM calls (\textbf{\#LLM Calls}), successful state transitions collected (\textbf{\#Transitions}), and language feedback provided (\textbf{\#Feedback}) across the three runs in each domain.}
\label{table:stats}
\end{table}

\subsubsection{Runtime Analysis}

To gain insights into the computational efficiency of \interpret, we measure its runtime in the simulated domains and provide a breakdown by stage in \cref{table:runtime}. Due to the variability in runtime across different iterations, we report the median, minimum, and maximum values for a comprehensive overview. The results reveal that the GPT-4-powered predicate learning and goal translation stages constitute the primary computational bottleneck. This is expected as calling the GPT-4 API involves a relatively long waiting time, which is also significantly influenced by the quality of the Internet connection. However, we anticipate that response time will cease to be a limiting factor for \acp{llm} in the near future, given the rapid advancements in the field.

We also present other relevant statistics in \cref{table:stats}, including the number of LLM calls, successful state transitions collected, and the amount of human feedback provided across three training runs for each domain. While these values exhibit considerable variation due to the inherent randomness in exploration and LLM outputs, \interpret demonstrates the ability to recover a PDDL domain from a relatively small number of language feedback and interaction data. This highlights the sample efficiency of our approach, which is crucial for practical applications where extensive human feedback and interaction may be costly or time-consuming to obtain.

\begin{table*}[ht]
\centering
{\scriptsize
\begin{tabular}{l|c@{}c@{}c@{}}
\toprule
\addlinespace
& \textbf{Run1} & \textbf{Run2} & \textbf{Run3} \\ 
\midrule
\textbf{Goal Predicates} & 
\begin{tabular}{@{}c@{}} 
\texttt{obj\_on\_obj(a, b)}, \\ 
\texttt{obj\_on\_table(a)} 
\end{tabular} & 
\begin{tabular}{@{}c@{}} 
\texttt{obj\_on\_obj(a, b)}, \\ 
\texttt{obj\_on\_table(a)} 
\end{tabular} & 
\begin{tabular}{c} 
\texttt{obj\_on\_obj(a, b)}, \\ 
\texttt{obj\_on\_table(a)} 
\end{tabular} \\
\addlinespace
\hline
\addlinespace
\textbf{Precondition Predicates} & 
\begin{tabular}{c} 
\texttt{obj\_small\_enough(a)}, \\ 
\texttt{obj\_clear(a)},\\ 
\texttt{gripper\_empty()} 
\end{tabular} & 
\begin{tabular}{c} 
\texttt{obj\_size\_ok\_for\_gripper(a)}, \\ 
\texttt{no\_obj\_on\_top(a)},\\ 
\texttt{hand\_empty()} 
\end{tabular} & 
\begin{tabular}{c} 
\texttt{obj\_small\_enough\_for\_gripper(a)}, \\ 
\texttt{obj\_free\_of\_objects(a)}, \\
\texttt{gripper\_empty()} 
\end{tabular} \\
\addlinespace
\bottomrule
\end{tabular}
\caption{\textbf{Learned predicates across three training runs with varied language feedback for the simulated StoreObjects domain.}
\label{table:diverse_pred}}
}
\end{table*}

\begin{table}[ht]
\centering
\small
\begin{tabular}{lccc}
\toprule
& Run1 & Run2 & Run3 \\ 
\midrule
\texttt{Canonical} & 1.00 & 1.00 & 1.00 \\
\texttt{More objects} & 1.00 & 1.00 & 1.00 \\
\texttt{Novel goals} & 1.00 & 1.00 & 1.00 \\
\texttt{Combined} & 1.00 & 1.00 & 1.00 \\
\bottomrule
\end{tabular}
\caption{\textbf{Evaluating \interpret trained with varied language feedback for the simulated StoreObjects domain. } We report the results of all three training runs. }
\label{table:diverse_results}
\end{table}

\subsubsection{Robustness to varied language feedback}

Natural language feedback from non-expert human users can be varied, with the same predicate being referred to in different ways. We evaluate the robustness of \interpret to such varied feedback in the simulated StoreObjects domain by synthesizing diverse feedback templates. Using ChatGPT~\cite{chat2023openai}, we generate 3 alternatives for each possible feedback, which are randomly sampled during each training step. We conduct three training runs with this varied feedback and perform both qualitative and quantitative evaluations.

\cref{table:diverse_pred} presents the predicates learned across the three runs, demonstrating that \interpret robustly captures the essential goal and precondition predicates. As the goal specifications are generally consistent, \interpret learns goal predicates with the same names in all runs. Despite the varied explanations of precondition violations, \interpret learns precondition predicates with different names but consistent semantics. \cref{table:diverse_results} shows that the predicates and operators learned from the varied feedback yield robust planning in all test sets. These results demonstrate \interpret's robustness to diverse language feedback, highlighting its ability to capture the underlying semantics despite variations in the feedback provided.

\section{Conclusions and Limitations}
\label{sec:conclude}

We present \interpret, an interactive framework that enables robots to learn symbolic predicates and operators from language feedback during embodied interaction. \interpret learns predicates as Python functions leveraging the capabilities of \acp{llm} like GPT-4. It allows iterative correction of these learned predicate functions based on human feedback to capture the core knowledge for planning. The predicates and operators learned by \interpret can be compiled on the fly as a \ac{pddl} domain, which enables effective task planning with a formal guarantee with a PDDL planner.
Our results demonstrate that \interpret can effectively acquire meaningful planning-oriented predicates, which allows learning operators to generalize to novel test tasks. In simulated domains, it achieves a 73\% success rate on the most challenging test set that requires generalizability to more objects and novel task goals. We also show \interpret can be applied in real-robot tasks. These findings validate our hypothesis that human-like planning proficiency requires interactive learning from rich language input, akin to infant development.

While showing promise, \interpret has several limitations that we would like to acknowledge. First, the generalizable planning capability of \interpret is realized by the learned symbolic operators. This introduces the assumption that the underlying domain can be well modeled at a symbolic level. This is generally not exact, as the physical world is inherently continuous. A promising future direction is to extend \interpret into the setup of \ac{tamp}~\cite{garrett2021integrated}, which considers both symbolic understanding and continuous interactions. Also, the operators learned by \interpret are deterministic, which falls short of capturing the uncertainty in state transitions. This issue can be mitigated by learning operators with probabilistic effects~\cite{konidaris2018skills,ahmetoglu2022deepsym}.

\bibliographystyle{unsrt}
\bibliography{main}

\onecolumn
\section*{APPENDIX A \\ Implementation Details}

\subsection*{A.I Operator Learning}
With a set of known predicates, operator learning from interaction data has been studied in Task and Motion Planning (TAMP)~\cite{silver2021learning,curtis2022discovering,kumar2023learning}. In this work, we implement a vairant of the Cluster-and-search operator learning algorithm~\cite{silver2021learning} to learn deterministic operators $\Omega$ from: (1) collected symbolic state transitions $\mathcal{D}_s=\{(s_{pre},\underline{a},s_{post})\}$ (where actions are successfully executed) and (2) action preconditions summarized from language feedback $\{\text{CON}_a\}$. Note that the symbolic transitions are pre-calculated by parsing the raw states $o_{pre}$ and $o_{post}$ with the learned predicates $\Psi$. We briefly describe how this algorithm works below.

Our operator learning algorithm aims to explain as many transitions in $\mathcal{D}_s$ with the learned operators $\Omega$, while preserving the known action preconditions $\{\text{CON}_a\}$ in $\Omega$. Formally, an operator $\omega$ is parameterized by a list of object variables $\text{PAR}$, while the lifted action $a$, precondition and effect sets $\text{CON}$, $\text{EFF}^+$ and $\text{EFF}^-$ are defined with respect to $\text{PAR}$. A symbolic transition $(s_{pre},\underline{a},s_{post})$ is explained by an operator $\omega$ when there exists an assignment from $\text{PAR}$ to specific objects in the scene, so that the grounded sets follow $\overline{\text{CON}} \in s_{pre}$, $\overline{\text{EFF}^-}=(s_{pre}-s_{post})$ and $\overline{\text{EFF}^+}=(s_{post}-s_{pre})$. We learn $\Omega$ to maximize the number of symbolic transitions explained while subject to the constraints below: (i) Each symbolic transition can be explained by at most one operator; (ii) No two operators have the same precondition sets but different effects; (iii) All operators corresponding to primitive action $a$ must have $\{\text{CON}_a\}$ in their preconditions; (iv) All operators keep a minimal set of preconditions as possible. In practice, learning such a set of operators can be achieved in two steps following~\cite{silver2021learning}: (i) initialize operators by clustering symbolic transitions by their lifted effects and preconditions, and (ii) searching for the final operators with minimal precondition set by keeping removing lifted preconditions and merging the initialized operators. We refer the readers to the original paper for more details.

\subsection*{A.2 Real-robot Setup}

We conduct real-robot experiments on a 7-Dof Franka Panda robot arm, where a table-mounted Kinect Azure camera provides RGB-D workspace observations. 

\paragraph*{Perception APIs} We implement a set of perception APIs to retrieve the semantic and geometric information of detected objects. We detect and segment tabletop objects with Grounded-SAM~\cite{ren2024grounded}, and use the object masks to crop the observed depth map to obtain object point clouds. Then we can estimate the geometric information, \eg, 3D object bounding boxes, based on the object point clouds. To , we use higher thresholds for detection and classification,  The full list of implemented perception APIs can be found in Appendix B.2.

\paragraph*{Primitive Actions} We implement heuristic-based action primitives to physically pick, place, and push objects with the robot arm. We utilize the Deoxys~\cite{zhu2022viola} library for low-level control.

\paragraph*{Handling Perception Uncertainty}

To enable robust predicate learning, we also implement various strategies to handle real-world perception errors. We observe two types of errors that could happen: (i) misdetected or misclassified objects, and (ii) inaccurately estimated 3D object bounding boxes due to noisy depth maps and occlusions. Our strategies include:
    \begin{itemize}
        \item Increase detection and classification thresholds to avoid false positives. We skip the predicate learning step when a task-relevant object is not detected.
        \item Filter out noisy outliers for point clouds to improve 3D object bounding box accuracy.
        \item Prompt GPT-4 to generate predicate functions that handle uncertainties. For example, testing equality between $x$ and $b$ is represented as $|x-y|<tol$, where $tol$ is a constant tolerance that is estimated and refined during iterative correction.
        \item Iteratively refine existing predicate functions to correct errors by prompting as the robot interacts with the environment and receives new observations and feedback.
\end{itemize}

\subsection*{A.3 Varied Feedback Experiment Setup}

In the varied feedback experiment, we synthesize three alternatives for each language feedback template with ChatGPT~\cite{chat2023openai}, which are used to evaluate the robustness of \interpret against varied language.
We present the synthesized feedback templates for the StoreObjects domain as below. 

\begin{itemize}
    \item Explain infeasible action
        \begin{itemize}
            \item Precondition 1: you can't execute $\texttt{pick\_up(a)}$ because
            \begin{itemize}
                \item the gripper is already occupied
                \item the gripper already held an object
                \item the gripper has an object in hand and can't pick up more objects
            \end{itemize}
            
            \item Precondition 2: you can't execute $\texttt{pick\_up(a)}$ because 
            \begin{itemize}
                \item there is something on $a$
                \item object $a$ has another object on its top
                \item there is another object on object $a$ so you can't pick it up
            \end{itemize}
            
            \item Precondition 3: you can't execute $\texttt{pick\_up(a)}$ because 
            \begin{itemize}
                \item $a$ can not be grasped by the gripper as it is too wide
                \item object $a$ is too large for the gripper to pick up
                \item you can't pick up object $a$ because it is too large
            \end{itemize}
            
            \item Precondition 4: you can't execute $\texttt{place\_on\_table(a)}$ because 
            \begin{itemize}
                \item object $a$ is not held by the gripper
                \item object the gripper does not hold object $a$
                \item object $a$ is not in the gripper
            \end{itemize}
            
            \item Precondition 5: you can't execute $\texttt{place\_first\_on\_second(a,b)}$ because 
            \begin{itemize}
                \item object $a$ is not held by the gripper
                \item object the gripper does not hold object $a$
                \item object $a$ is not in the gripper
            \end{itemize}
            
            \item Precondition 6: you can't execute $\texttt{place\_first\_on\_second(a,b)}$ because 
            \begin{itemize}
                \item there is something on $b$
                \item object $b$ has another object on its top
                \item there is another object on object $b$ so you can't pick it up
            \end{itemize}
        \end{itemize}
    \item Explain unsatisfied goal conditions
        \begin{itemize}
            \item You haven't achieved the goal because 
            \begin{itemize}
                \item object $a$ is not yet on $b$
                \item object $a$  has not been put on object $b$ 
                \item you haven't put object $a$ on object $b$
            \end{itemize}
            
            \item You haven't achieved the goal because 
            \begin{itemize}
                \item object $a$ is not on table
                \item object $a$ is not yet on table
                \item you haven't put object $a$ on table
            \end{itemize}
        \end{itemize}
\end{itemize}

\section*{APPENDIX B \\ Full Prompt Templates}

\subsection*{B.1 Reasoner}

We present the prompt templates of \textit{Reasoner} for different types of language feedback below.

\vspace{2mm}

\subsubsection*{1) Specify goal}

\begin{codeframe}
\footnotesize
\begin{lstlisting}[style=RawText, captionpos=b]
You are a helpful assistant that converts natural language goals into symbolic goals, by proposing necessary predicates to learn. A predicate is a function that takes objects as arguments and outputs True or False. Please reason about predicates that directly reflect the goals, and only include these predicates in the symbolic goal. Please try to reuse available predicates to avoid redundant predicates. When give the symbolic goal, only include the symbolic literals that are directly mentioned in the goal text. The literals only take available objects as arguments (not "table", "any", or other variables). Do not infer or guess on what other literals should be in the goal. If you can't come up with goal literals given the above constraints, think about inventing new predicates. Other known environment entities include "table" and "gripper", so you can invent predicates such as "in_gripper", "under_table", etc.
Note that the predicates in the given example are unknown to you.

Example:
Available objects: ['plate']
Available predicates: {'obj_in_sink(a)': 'check whether object a is in sink or not'}
Goal: Wash the plate and move it out of the sink.
Output:
{
    "Reasoning": "The goal directly captures two symbolic literals, obj_washed(plate) and obj_in_sink(plate). As predicate obj_in_sink(a) is available, we only need to invent predicate obj_washed(a).",
    "Invented predicates": {
        "obj_washed(a)": "check whether object a is washed or not"
    },
    "Symbolic goal": {
        "obj_washed(plate)": true,
        "obj_in_sink(plate)": false
    }
}

{domain_desc}
Available objects: {entities}
Available predicates: {predicates}
Goal: {goal_spec}
Please give the output following the format in the examples, and make sure to include all fields of the json in your outputs.
Output (please don't output ```json):
\end{lstlisting}
\end{codeframe}

\subsubsection*{2) Explain unsatisfied goal}

\begin{codeframe}
\footnotesize
\begin{lstlisting}[style=RawText, captionpos=b]
You are a helpful assistant that translates human explanations of the current state into symbolic literals, using the available predicates. The human explanations are about why the goal is not achieved in the current state.

Example:
Available objects: ['cup', 'block', 'plate']
Available predicates: {'holding(x)': 'check whether the gripper is holding x or not', 'in_sink(x)': 'check whether x is in sink or not'}
Human explanation: The cup is not in the sink, and it is held by the gripper.
Output:
{
    "Current state": {
        "holding(cup)": true,
        "in_sink(cup)": false
    }
}

{domain_desc}
Available objects: {entities}
Available predicates: {predicates}
Human explanation: {human_explain}
Output (please don't output ```json):
\end{lstlisting}
\end{codeframe}

\subsubsection*{3) Explain infeasible action}

\begin{codeframe}
\footnotesize
\begin{lstlisting}[style=RawText, captionpos=b]
You are a helpful assistant that invents predicates to describe action preconditions, based on human explanations of why an action is infeasible. A predicate is a function that takes objects as arguments and outputs True or False. A literal (or an atom) grounds a predicate on available objects (but not "table", "any", or other variables).
Please generate output step-by-step:
1. Reasoning: Based on the human explanation, reason about the current state and the preconditions of executing the action. Then invent predicates that directly represent the action preconditions. The predicates can only take in object variables as inputs, but not environment-related entities such as "gripper", "robot", or vague entities such as "any object". Predicates can have empty arguments. Please try to reuse available predicates to avoid inventing redundant predicates.
2. Invented predicates: Based on the reasoning in Step 1, list the invented predicates and their explanations. Make sure to include as detailed explanations as possible according to the human explanation. The predicates only take object variables as arguments (not numerical variables).
3. New action preconditions: Based on the reasoning in Step 1, give the new preconditions of the action that take lifted variables such as "a", "b" as arguments. Please do not include other preconditions that are not mentioned in the human explanation.
4. Current state literals: Based on the explanation and reasoning in Step 1, give the curent state literals mentioned in the human explanation, using the invented and existing predicates. Please do not include other literals that are not mentioned in the human explanation. If you are not sure about whether to include a literal, don't include it.

Note that the predicates in the given example are unknown to you.

Example:
Available objects: ['cup', 'plate']
Available predicates: {}
Infeasible action: wash(plate)
Human explanation: You can't wash plate because the robot is far away from the plate, the robot needs to be within 1m from the plate to wash it; also, the robot has something else in hand.
Output:
{
    "1. Reasoning": "Based on human explanation, we know the robot is farther than 1m from the plate, and it has something else in hand, so it can't to wash the plate. The precondition of action wash(plate) is that the robot is close enough (smaller than 1m) to the plate, and its hand is empty. Given the available predicates, none of them can directly represent the precondition; so we invent predicate obj_close_enough(a) to check whether object a is close enough, and predicate hand_empty() to check whether robot has its hand free. Then the current literal follows obj_close_enough(plate) is false and hand_empty() is false, and the action precondition is obj_close_enough(plate) is true and hand_empty() is true."
    "2. Invented predicates": {
        "obj_close_enough(a)": "check whether object a is close enough to the robot, the predicate holds when the distance between robot and object is smaller that 1m",
        "hand_empty()": "check whether the robot has its hand free, the predicate holds when the robot is not holding anything"
    },
    "3. New action preconditions": {
        "action": "wash(a)",
        "new preconditions": {
            "obj_close_enough(a)": true,
            "hand_empty()": true
        }
    },
    "4. Current state literals": {
        "obj_close_enough(plate)": false,
        "hand_empty()": false
    }
}

{domain_desc}
Available objects: {entities}
Available predicates: {predicates}
Infeasible action: {failed_action}
Human explanation: {failure_explain}
Please give the output following the format in the examples, and make sure to include all fields of the json in your outputs.
Output (please don't output ```json):
\end{lstlisting}
\end{codeframe}

\subsection*{B.2 Coder}

\vspace{2mm}

\subsubsection*{1) Main template}

\begin{codeframe}
\footnotesize
\begin{lstlisting}[style=RawText, captionpos=b]
You are a helpful assistant that writes python functions to ground the given predicates. A set of perception API functions are available to provide the basic perception information.
You can write extra utility functions that can be used in your predicate functions. You can also introduce new dependent predicates when necessary, but please keep the set of predicates as compact as possible. Note that some predicates may depend on each other; so you can reuse predicate functions in other predicates to avoid writing duplicate code.
{domain_desc} Please find the available API functions, and examples of utility and predicate functions below.

{known_functions}

The observation is: {observation}

Now you are asked to ground the predicates below: {new_predicates}
Please put "# <predicate>" and "# <end-of-predicate>", "# <utility>" and "# <end-of-utility>" at the beginning and end of each predicate function and utility function. Remember to include description surrounded with "<<", ">>" for predicates.
\end{lstlisting}
\end{codeframe}

\subsubsection*{2) Code example for simulated domains}

\begin{codeframe}
\footnotesize
\begin{lstlisting}[style=PythonStyle, captionpos=b]
import numpy as np
from typing import *
from predicate_learning.predicate_gym.perception_api import (
    get_detected_object_list,
    get_object_xy_position,
    get_object_xy_size,
    get_object_category,
    get_object_water_amount,
    get_gripper_position,
    get_gripper_open_width,
    get_in_gripper_mass,
    get_gripper_y_size,
    get_gripper_max_open_width,
    get_table_x_range,
    get_table_y_height,
)

eps = 0.2

# Utility functions:
# <utility>
def get_object_xy_bbox(a) -> np.ndarray:
    """
    Get the xyxy bounding box of object a
    :param a: string, name of detected object
    :return: np.ndarray, [x1, y1, x2, y2], where x1 is left, x2 is right, y1 is bottom, y2 is top
    """
    object_a_position = get_object_xy_position(a)
    object_a_size = get_object_xy_size(a)
    return [
        object_a_position[0] - object_a_size[0] / 2,
        object_a_position[1] - object_a_size[1] / 2,
        object_a_position[0] + object_a_size[0] / 2,
        object_a_position[1] + object_a_size[1] / 2,
    ]
# <end-of-utility>

# <utility>
def get_in_gripper_xy_bbox() -> np.ndarray:
    """
    Get the xyxy bounding box of space within the gripper
    :param a: string, name of detected object
    :return: np.ndarray, [x1, y1, x2, y2], where x1 is left, x2 is right, y1 is bottom, y2 is top
    """
    gripper_position = get_gripper_position()
    gripper_open_width = get_gripper_open_width()
    gripper_y_size = get_gripper_y_size()
    return [
        gripper_position[0] - gripper_open_width / 2,
        gripper_position[1] - gripper_y_size / 2,
        gripper_position[0] + gripper_open_width / 2,
        gripper_position[1] + gripper_y_size / 2,
    ]
# <end-of-utility>

# Predicates:
# <predicate>
def obj_in_gripper(a) -> bool:
    """
    Description: <<whether object a is held by the gripper>>
    The predicate holds True when the mass in gripper is non-zero, object a is aligned with the opened gripper along x axis, and overlaps with the gripper along y axis
    :param a: string, name of detected object
    :return: bool
    """
    # get in gripper mass
    in_gripper_mass = get_in_gripper_mass()
    # get in_gripper xyxy bbox
    in_gripper_xyxy_bbox = get_in_gripper_xy_bbox()

    # get bbox_xyxy of object a
    object_a_xyxy_bbox = get_object_xy_bbox(a)

    # check whether the mass in gripper is non-zero
    # in_gripper_mass > eps
    if in_gripper_mass > eps:
        # check whether the object a is aligned with the gripper along x axis
        # abs(a.x1 - gripper.x1) < eps and abs(a.x2 - gripper.x2) < eps
        if (
            np.abs(object_a_xyxy_bbox[0] - in_gripper_xyxy_bbox[0]) < eps
            and np.abs(object_a_xyxy_bbox[2] - in_gripper_xyxy_bbox[2]) < eps
        ):
            # check whether the object a overlaps with the gripper along y axis
            # a.y1 < gripper.y2 + eps and a.y2 > gripper.y1 - eps
            if (
                object_a_xyxy_bbox[1] < in_gripper_xyxy_bbox[3] + eps
                and object_a_xyxy_bbox[3] > in_gripper_xyxy_bbox[1] - eps
            ):
                return True
            else:
                return False
        else:
            return False
    else:
        return False
# <end-of-predicate>
\end{lstlisting}
\end{codeframe}

\subsubsection*{3) Code example for real-world domains}

\begin{codeframe}
\footnotesize
\begin{lstlisting}[style=PythonStyle, captionpos=b]
import numpy as np
from typing import *
from real_robot.perception_api import (
    get_detected_object_list,
    get_object_category,
    get_object_center_3d,
    get_object_size_3d,
    get_gripper_position_3d,
    get_gripper_max_open_width,
    get_gripper_open_width,
    get_table_height,
)

# All numbers are in meters. Please reason about the proper tolerance value for each predicate to incoporate real-world perception uncertainty. You may design different tolerance for different variables, like along different axis.
# Usually, when objects are close to each other (e.g., stacked together), the segmented objects may include part of each other and their bounding boxes may overlap. You need to handle this using the tolerance properly.

# Utility functions:
# <utility>
def get_object_x_range(a) -> np.ndarray:
    """
    Get the range of object a along x axis
    :param a: string, name of detected object
    """
    center = get_object_center_3d(a)
    extent = get_object_size_3d(a)
    return np.array([center[0] - extent[0] / 2, center[0] + extent[0] / 2])
# <end-of-utility>

# Predicates:
# <predicate>
def obj_in_gripper(a) -> bool:
    """
    Description: <<whether object a is held by the gripper>>
    The predicate holds True when gripper is half open and object a is close to the gripper.
    :param a: string, name of detected object
    """
    gripper_open_tol = 0.01
    z_tol = 0.01
    xy_tol = 0.01
    
    # check whether gripper is half-open
    # gripper_width < max_gripper_width + gripper_open_tol
    if get_gripper_open_width() < get_gripper_max_open_width() - gripper_open_tol:
        gripper_position = get_gripper_position_3d()
        object_center = get_object_center_3d(a)
        object_extent = get_object_size_3d(a)
        # check whether the distance between object a and gripper along z is within z_tol
        # abs(gripper_z - object_center_z) < z_tol
        if np.abs(gripper_position[2] - object_center[2]) < z_tol:
            # check whether the distance between object a and gripper along x and y is within half the extent of a
            # abs(gripper_xy - object_center_xy) < object_extent_xy / 2 + xy_tol
            if np.all(np.abs(gripper_position[:2] - object_center[:2]) < object_extent[:2] / 2 + xy_tol):
                return True
            else:
                False
        else:
            False
    else:
        return False
# <end-of-predicate>
\end{lstlisting}
\end{codeframe}

\subsection*{B.3 Corrector}
\setcounter{subsubsection}{0}

\vspace{2mm}

\subsubsection*{1) Execution error}

\begin{codeframe}
\footnotesize
\begin{lstlisting}[style=RawText, captionpos=b]
You are a helpful assistant that modifies the predicate grounding functions based on the execution error.

{domain_desc}

{known_functions}

The observation is: {observation}

The execution error is: {error}

The error trace is: {trace}

Please answer the two questions below:
1. What is your reasoning for the execution error? How would you modify the predicate grounding functions to fix the error?
2. Based on your reasoning, please return the modified functions without further explanations. Please put "# <predicate>" and "# <end-of-predicate>", "# <utility>" and "# <end-of-utility>" at the beginning and end of each predicate function and utility function, following the format of the original functions. Remember to include description surrounded with "<<", ">>" for predicates.

\end{lstlisting}
\end{codeframe}

\subsubsection*{2) Alignment error}

\begin{codeframe}
\footnotesize
\begin{lstlisting}[style=RawText, captionpos=b]
You are a helpful assistant that modifies the predicate functions based on correction from human. You can introduce new utility functions and predicates in your modification when necessary. Note that the inconsistency might not due to the direct implementation of the predicate, but other predicate functions or utility functions it depends on. If this is the case, you need to modify its dependent predicates andutility functions.
{domain_desc} The current predicate functions are below.

{known_functions}

The observation is: {observation}

The human correction is: {correction}

Please answer the two questions below:
1. What is your reasoning for the human corection? Which part of the predicate functions is wrong? How would you modify the predicate grounding functions to reflect the correction?
2. Based on your reasoning, please return the modified functions without further explanations. Please put "# <predicate>" and "# <end-of-predicate>", "# <utility>" and "# <end-of-utility>" at the beginning and end of each predicate function and utility function, following the format of the original functions. Remember to include description surrounded with "<<", ">>" for predicates.
\end{lstlisting}
\end{codeframe}

\subsection*{B.4 Goal Translator}
\vspace{2mm}

\begin{codeframe}
\footnotesize
\begin{lstlisting}[style=RawText, captionpos=b]
You are a helpful assistant that converts natural language goals into symbolic goals. The symbolic goal must only use available predicates.
It is fine if the natural language goal can not be fully captured by the available predicates, please just output the symbolic goal that is as close as possible to the natural language goal.

Example:
Available entities: ['plate']
Available predicates: {'obj_in_sink(a)': 'check whether object a is in sink or not'}
Goal: Wash the plate and put it in sink.
Output:
{
    "Symbolic goal": ["obj_in_sink(plate)"]
}

{domain_desc}
Available entities: {entities}
Available predicates: {predicates}
Goal: {goal_spec}
Output (please don't output ```json):
\end{lstlisting}
\end{codeframe}

\section*{APPENDIX C \\ Domain Design}

We provide further details of the designed domains including: (i) the available objects, (ii) the available primitive actions, (ii) Simple and complex tasks, and (iii) language feedback templates.

\subsection*{C.1 StoreObjects}
\paragraph*{Available objects}
\begin{itemize}
    \item A large object, \ie, a shelf or coaster that can not be picked up
    \item A number of small objects that are to be stored on the large object
\end{itemize}

\paragraph*{Available primitive actions}
\begin{itemize}
    \item $\texttt{pick\_up(a)}$: pick up an object $a$
    \item $\texttt{place\_on\_table(a)}$: place an object $a$ on table
    \item $\texttt{place\_first\_on\_second(a,b)}$: place an object $a$ on object $b$
\end{itemize}

\paragraph*{Simple and complex tasks}
\begin{itemize}
    \item Simple task 1: stack a small object $a$ on the large object $b$
    \item Simple task 2: stack a small object $a$ on a small object $b$
    \item Simple task 3: put a small object $a$ on table
    \item Complex task: store objects on the large object $a$ following the order: $b$ on $a$, $c$ on $b$, ...
\end{itemize}

\paragraph*{Language feedback templates}
\begin{itemize}
    \item Explain infeasible action
        \begin{itemize}
            \item Precondition 1: you can't execute $\texttt{pick\_up(a)}$ because the gripper is already occupied
            \item Precondition 2: you can't execute $\texttt{pick\_up(a)}$ because there is something on $a$
            \item Precondition 3: you can't execute $\texttt{pick\_up(a)}$ because $a$ can not be grasped by the gripper as it is too wide
            \item Precondition 4: you can't execute $\texttt{place\_on\_table(a)}$ because object $a$ is not held by the gripper
            \item Precondition 5: you can't execute $\texttt{place\_first\_on\_second(a,b)}$ because object $a$ is not held by the gripper
            \item Precondition 6: you can't execute $\texttt{place\_first\_on\_second(a,b)}$ because there is something on $b$
        \end{itemize}
    \item Explain unsatisfied goal
        \begin{itemize}
            \item Unsatisfied goal 1: you haven't achieved the goal because object $a$ is not yet on $b$
            \item Unsatisfied goal 2: you haven't achieved the goal because object $a$ is not on table
        \end{itemize}
\end{itemize}

\subsection*{C.2 SetTable}
\paragraph*{Available objects}
\begin{itemize}
    \item A table mat that can not be moved
    \item A plate that can only be pushed but not grasped
    \item A number of small objects that are to be placed on plate / table mat
\end{itemize}

\paragraph*{Available primitive actions}
\begin{itemize}
    \item $\texttt{pick\_up(a)}$: pick up an object $a$
    \item $\texttt{place\_on\_table(a)}$: place an object $a$ on table
    \item $\texttt{place\_first\_on\_second(a,b)}$: place an object $a$ on object $b$
    \item $\texttt{push\_plate\_on\_object(a,b)}$: push a plate $a$ onto object $b$
\end{itemize}

\paragraph*{Simple and complex tasks}
\begin{itemize}
    \item Simple task 1: place a small object $a$ on table mat / plate $b$
    \item Simple task 2: place the plate $a$ on table mat $b$
    \item Simple task 3: place a small object $a$ on table
    \item Complex task: set a breakfast table with plate $b$ on table mat $a$, $c$ on $b$, ...
\end{itemize}

\paragraph*{Language feedback templates}
\begin{itemize}
    \item Explain infeasible action
        \begin{itemize}
            \item The 6 precondition explanations as in StoreObjects
            \item Precondition 7: you can't execute $\texttt{push\_plate\_on\_object(a,b)}$ because object $a$ is not a plate
            \item Precondition 8: you can't execute $\texttt{push\_plate\_on\_object(a,b)}$ because the gripper is occupied
            \item Precondition 9: you can't execute $\texttt{push\_plate\_on\_object(a,b)}$ because there is something on $a$
            \item Precondition 10: you can't execute $\texttt{push\_plate\_on\_object(a,b)}$ because there is something on $b$
            \item Precondition 11: you can't execute $\texttt{push\_plate\_on\_object(a,b)}$ because object $b$ is not thin enough as its height is greater than xxx
        \end{itemize}
    \item Explain unsatisfied goal
        \begin{itemize}
            \item Unsatisfied goal 1: you haven't achieved the goal because object $a$ is not yet on $b$
            \item Unsatisfied goal 1: you haven't achieved the goal because object $a$ is not on table
        \end{itemize}
\end{itemize}

\subsection*{C.3 CookMeal}
\paragraph*{Available objects}
\begin{itemize}
    \item A heavy pot that can contain food ingredients and water, but can not be moved by the gripper
    \item A faucet that can get water from with a container
    \item One or more cups that can be used to get water from facet and contain water
    \item A number of food ingredients that are to be put into the pot
\end{itemize}

\paragraph*{Available primitive actions} 
\begin{itemize}
    \item $\texttt{pick\_up(a)}$: pick up an object $a$
    \item $\texttt{place\_on\_table(a)}$: place an object $a$ on table
    \item $\texttt{place\_first\_in\_second(a,b)}$: place an object $a$ into container $b$
    \item $\texttt{get\_water\_from\_faucet(a)}$: get water from the faucet with cup $a$
    \item $\texttt{pour\_water\_from\_first\_to\_second(a,b)}$: pour water from container $a$ to container $b$
\end{itemize}

\paragraph*{Simple and complex tasks}
\begin{itemize}
    \item Simple task 1: pour water into cup $a$ and put it on table
    \item Simple task 2: pour water into pot $a$
    \item Simple task 3: put object $a$ into pot $b$
    \item Complex task: pour water and put $a$, $b$, ... in pot $c$, pour water into cup $d$ and put it on table
\end{itemize}

\paragraph*{Language feedback templates}
\begin{itemize}
    \item Explain infeasible action
        \begin{itemize}
            \item Precondition 1: you can't execute $\texttt{pick\_up(a)}$ because the gripper is already occupied
            \item Precondition 2: you can't execute $\texttt{pick\_up(a)}$ because $a$ is in a container
            \item Precondition 3: you can't execute $\texttt{pick\_up(a)}$ because $a$ can not be grasped by the gripper as it is too wide
            \item Precondition 4: you can't execute $\texttt{place\_on\_table(a)}$ because object $a$ is not held by the gripper
            \item Precondition 5: you can't execute $\texttt{place\_first\_in\_second(a,b)}$ because object $a$ is not held by the gripper
            \item Precondition 6: you can't execute
            $\texttt{place\_first\_in\_second(a,b)}$ because object $b$ is not a container, only objects with category cup, pot, and basket are containers
            \item Precondition 7: you can't execute $\texttt{place\_first\_in\_second(a,b)}$ because object $a$ is not food
            \item Precondition 8: you can't execute $\texttt{place\_first\_in\_second(a,b)}$ because object $b$ can not contain food as it's too small, i.e., its width is smaller than 10
            \item Precondition 9: you can't execute $\texttt{get\_water\_from\_faucet(a)}$ because object $a$ is not held by the gripper
            \item Precondition 10: you can't execute $\texttt{get\_water\_from\_faucet(a)}$ because object $a$ is not a container
            \item Precondition 11: you can't execute $\texttt{get\_water\_from\_faucet(a)}$ because object $a$ already contains water
            \item Precondition 12: you can't execute $\texttt{pour\_water\_from\_first\_to\_second(a,b)}$ because object $a$ is not held by the gripper
            \item Precondition 13: you can't execute $\texttt{pour\_water\_from\_first\_to\_second(a,b)}$ because object $a$ does not have water
            \item Precondition 14: you can't execute $\texttt{pour\_water\_from\_first\_to\_second(a,b)}$ because object $b$ is not a container
        \end{itemize}
    \item Explain unsatisfied goal
        \begin{itemize}
            \item Unsatisfied goal 1: you haven't achieved the goal because object $a$ does not have water
            \item Unsatisfied goal 2: you haven't achieved the goal because object $a$ is not on table
            \item Unsatisfied goal 3: you haven't achieved the goal because object $a$ is not in pot $b$
        \end{itemize}
\end{itemize}

\section*{APPENDIX D \\ Examples of Learned Predicates and Operators}

We present examples of learned predicates and operators in the real-world SetTable domain. Note that these predicates and operators are pretty much a superset of those in the real-world StoreObjects domain. We show the predicates as Python functions and the operators in a PDDL domain file. For the learned predicates and operators in the simulated domains, please refer to our \href{https://github.com/hmz-15/Interactive-Predicate-Learning}{github repo}.

\begin{codeframe}
\footnotesize
\begin{lstlisting}[style=PythonStyle, caption={Learned Predicates for Real-world SetTable Domain}, captionpos=b]
import numpy as np
from typing import *
from real_robot.perception_api import (
    get_detected_object_list,
    get_object_category,
    get_object_center_3d,
    get_object_size_3d,
    get_gripper_position_3d,
    get_gripper_max_open_width,
    get_gripper_open_width,
    get_table_height,
)

# All numbers are in meters. Please reason about the proper tolerance value for each predicate to incoporate real-world perception uncertainty. You may design different tolerance for different variables, like along different axis.
# Usually, when objects are close to each other (e.g., stacked together), the segmented objects may include part of each other and their bounding boxes may overlap. You need to handle this using the tolerance properly.

# Utility functions:
# <utility>
def get_object_x_range(a) -> np.ndarray:
    """
    Get the range of object a along x axis
    :param a: string, name of detected object
    """
    center = get_object_center_3d(a)
    extent = get_object_size_3d(a)
    return np.array([center[0] - extent[0] / 2, center[0] + extent[0] / 2])
# <end-of-utility>

# <utility>
def get_object_y_range(a) -> np.ndarray:
    """
    Get the range of object a along y axis
    :param a: string, name of detected object
    """
    center = get_object_center_3d(a)
    extent = get_object_size_3d(a)
    return np.array([center[1] - extent[1] / 2, center[1] + extent[1] / 2])
# <end-of-utility>

# Predicates:
# <predicate>
def obj_in_gripper(a) -> bool:
    """
    Description: <<whether object a is held by the gripper>>
    The predicate holds True when gripper is half open and object a is close to the gripper.
    :param a: string, name of detected object
    """
    gripper_open_tol = 0.01
    z_tol = 0.02
    xy_tol = 0.01
    
    # check whether gripper is half-open
    # gripper_width < max_gripper_width + gripper_open_tol
    if get_gripper_open_width() < get_gripper_max_open_width() - gripper_open_tol:
        gripper_position = get_gripper_position_3d()
        object_center = get_object_center_3d(a)
        object_extent = get_object_size_3d(a)
        # check whether the distance between object a and gripper along z is within z_tol
        # abs(gripper_z - object_center_z) < z_tol
        if np.abs(gripper_position[2] - object_center[2]) < z_tol:
            # check whether the distance between object a and gripper along x and y is within half the extent of a
            # abs(gripper_xy - object_center_xy) < object_extent_xy / 2 + xy_tol
            if np.all(np.abs(gripper_position[:2] - object_center[:2]) < object_extent[:2] / 2 + xy_tol):
                return True
            else:
                False
        else:
            False
    else:
        return False
# <end-of-predicate>

# <predicate>
def obj_on_obj(a: str, b: str) -> bool:
    """
    Description: <<check whether object a is on top of object b or not>>
    The predicate holds True if the bottom of object a is within a certain tolerance above the top of object b,
    and their x and y projections overlap.
    :param a: string, name of detected object
    :param b: string, name of detected object
    """
    z_tol = 0.01  # Reduced tolerance for the z-axis to consider object a is on top of object b
    overlap_tol = 0.01  # Reduced tolerance for the overlap in x and y axis
    
    # Get the center and size of both objects
    center_a, size_a = get_object_center_3d(a), get_object_size_3d(a)
    center_b, size_b = get_object_center_3d(b), get_object_size_3d(b)
    
    # Calculate the z position of the bottom of object a and the top of object b
    bottom_a = center_a[2] - size_a[2] / 2
    top_b = center_b[2] + size_b[2] / 2
    
    # Check if bottom of a is within tolerance above top of b
    if not (bottom_a <= top_b + z_tol and bottom_a >= top_b - z_tol):
        return False
    
    # Check if the projections of a and b on the x and y axes overlap
    x_range_a = get_object_x_range(a)
    x_range_b = get_object_x_range(b)
    y_range_a = get_object_y_range(a)
    y_range_b = get_object_y_range(b)
    
    # Check overlap in x and y axis
    overlap_x = min(x_range_a[1], x_range_b[1]) - max(x_range_a[0], x_range_b[0]) > -overlap_tol
    overlap_y = min(y_range_a[1], y_range_b[1]) - max(y_range_a[0], y_range_b[0]) > -overlap_tol
    
    return overlap_x and overlap_y
# <end-of-predicate>

# <predicate>
def gripper_empty() -> bool:
    """
    Description: <<check whether the gripper is empty, the predicate holds when the gripper is not holding any object>>
    """
    gripper_open_tol = 0.01  # Tolerance for considering the gripper as not fully open
    max_gripper_width = get_gripper_max_open_width()
    current_gripper_width = get_gripper_open_width()
    
    # If the gripper is almost fully open, we consider it empty
    if current_gripper_width >= max_gripper_width - gripper_open_tol:
        return True
    
    # If the gripper is not fully open, check proximity of objects to gripper
    detected_objects = get_detected_object_list()
    for obj in detected_objects:
        if obj_in_gripper(obj):
            return False  # Found an object in the gripper
    
    return True  # No object found in the gripper
# <end-of-predicate>

# <predicate>
def obj_on_table(a: str) -> bool:
    """
    Description: <<check whether object a is on the table or not>>
    The predicate holds True if the bottom of object a is within a certain tolerance above the table height.
    :param a: string, name of detected object
    """
    z_tol = 0.01  # Tolerance for considering object a is on the table

    # If object a is on another object, we consider it not on table
    detected_objects = get_detected_object_list()
    for obj in detected_objects:
        if obj == a:
            continue # Skip the object itself
        if obj_on_obj(a, obj):
            return False # Found object a on top of obj
    
    # Get the table height
    table_height = get_table_height()
    
    # Get the center and size of object a
    center_a, size_a = get_object_center_3d(a), get_object_size_3d(a)
    
    # Calculate the z position of the bottom of object a
    bottom_a = center_a[2] - size_a[2] / 2
    
    # Check if bottom of a is within tolerance above the table height
    return (bottom_a <= table_height + z_tol) and (bottom_a >= table_height - z_tol)
# <end-of-predicate>

# <predicate>
def obj_too_large_to_grasp(a: str) -> bool:
    """
    Description: <<check whether object a is too large for the gripper to grasp, the predicate holds when the object's size exceeds the gripper's grasping capacity>>
    :param a: string, name of detected object
    """
    # Get the maximum open width of the gripper
    max_gripper_width = get_gripper_max_open_width()
    
    # Get the size of object a
    size_a = get_object_size_3d(a)
    
    # Check if the object's size along the x axis (gripper open direction) exceeds the gripper's grasping capacity
    return size_a[0] > max_gripper_width
# <end-of-predicate>

# <predicate>
def obj_free_of_objects(a: str) -> bool:
    """
    Description: <<check whether object a does not have any other objects on top of it, the predicate holds when there are no objects placed on object a>>
    :param a: string, name of detected object
    """
    detected_objects = get_detected_object_list()
    for obj in detected_objects:
        if obj == a:
            continue  # Skip the object itself
        if obj_on_obj(obj, a):
            return False  # Found an object on top of object a
    return True  # No objects found on top of object a
# <end-of-predicate>

# <predicate>
def is_plate(a) -> bool:
    """
    Description: <<check whether object a is a plate, the predicate holds true if a is a plate>>
    :param a: string, name of detected object a
    """
    # Get the category of object a
    object_a_category = get_object_category(a)

    # Check whether the category of object a is plate
    if object_a_category == "plate":
        return True
    else:
        return False
# <end-of-predicate>

# <predicate>
def obj_thin_enough(a) -> bool:
    """
    Description: <<check whether object a is thin enough, the predicate holds true if the height of object a is less than or equal to 0.01>>
    :param a: string, name of detected object a
    """
    # Get the size of object a
    size_a = get_object_size_3d(a)

    # check whether the height of object a is less than or equal to 0.01
    if size_a[2] <= 0.01:
        return True
    else:
        return False
# <end-of-predicate>
\end{lstlisting}
\end{codeframe}

\begin{codeframe}
\footnotesize
\begin{lstlisting}[style=PythonStyle, caption={Learned Operators for Real-world SetTable Domain}, captionpos=b]

(define (domain set_table)
  (:requirements :typing)
  (:types default)
  
  (:predicates (obj_in_gripper ?v0 - default)
	(not_obj_in_gripper ?v0 - default)
	(obj_on_obj ?v0 - default ?v1 - default)
	(not_obj_on_obj ?v0 - default ?v1 - default)
	(gripper_empty)
	(not_gripper_empty)
	(obj_on_table ?v0 - default)
	(not_obj_on_table ?v0 - default)
	(obj_too_large_to_grasp ?v0 - default)
	(not_obj_too_large_to_grasp ?v0 - default)
	(obj_free_of_objects ?v0 - default)
	(not_obj_free_of_objects ?v0 - default)
        (is_plate ? v0 - default)
        (not_is_plate ? v0 - default)
        (obj_thin_enough ? v0 - default)
        (not_obj_thin_enough ? v0 - default)
	(pick_up ?v0 - default)
	(place_on_table ?v0 - default)
	(place_first_on_second ?v0 - default ?v1 - default)
  )
  ; (:actions pick_up place_on_table place_first_on_second push_plate_on_object)

	(:action place_on_table_1
		:parameters (?v_0 - default)
		:precondition (and (obj_in_gripper ?v_0)
			(place_on_table ?v_0))
		:effect (and
			(not_obj_in_gripper ?v_0)
			(gripper_empty)
			(not (obj_in_gripper ?v_0))
			(obj_on_table ?v_0))
	)

	(:action pick_up_1
		:parameters (?v_1 - default ?v_0 - default)
		:precondition (and (gripper_empty)
			(obj_free_of_objects ?v_0)
			(not_obj_too_large_to_grasp ?v_0)
			(pick_up ?v_0)
			(obj_on_obj ?v0 ?v_1))
		:effect (and
			(obj_in_gripper ?v_0)
			(not (obj_on_obj ?v_0 ?v_1))
			(obj_free_of_objects ?v_1)
			(not (gripper_empty))
			(not (not_obj_in_gripper ?v_0)))
	)

	(:action pick_up_2
		:parameters (?v_0 - default)
		:precondition (and (obj_on_table ?v_0)
			(gripper_empty)
			(obj_free_of_objects ?v_0)
			(not_obj_too_large_to_grasp ?v_0)
			(pick_up ?v_0))
		:effect (and
			(not (gripper_empty))
			(not (not_obj_in_gripper ?v_0))
			(obj_in_gripper ?v_0)
			(not (obj_on_table ?v_0)))
	)

	(:action place_first_on_second_1
		:parameters (?v_1 - default ?v_0 - default)
		:precondition (and (obj_in_gripper ?v_0)
			(obj_free_of_objects ?v_1)
			(place_first_on_second ?v_0 ?v_1))
		:effect (and
			(not_obj_in_gripper ?v_0)
			(not (obj_free_of_objects ?v_1))
			(gripper_empty)
			(obj_on_obj ?v_0 ?v_1)
			(not (obj_in_gripper ?v_0)))
	)

        (:action push_plate_on_object_1
		:parameters (?v_1 - default ?v_0 - default)
		:precondition (and (gripper_empty)
			(obj_free_of_objects ?v_0)
			(obj_free_of_objects ?v_1))
  			(obj_thin_enough ?v_1)
  			(is_plate ?v_0)
			(push_plate_on_object ?v_0 ?v_1))
		:effect (and
			(obj_on_obj ?v_0 ?v_1)
			(not (obj_free_of_objects ?v_1)))
	)

)
        
\end{lstlisting}
\end{codeframe}

\end{document}

%% file: configs.tex
\usepackage{graphicx} \graphicspath{ {figures/} }
\usepackage{amsmath,amssymb,mathabx}
\usepackage{algorithmic}
\usepackage[ruled,linesnumbered]{algorithm2e}
\usepackage{acronym}

\let\labelindent\relax
\usepackage{enumitem}
\usepackage{booktabs}
\usepackage[bookmarks=true,breaklinks,colorlinks,citecolor=purple]{hyperref}
\usepackage{bm}
\usepackage{balance}
\usepackage{xspace,setspace}
\usepackage[skip=3pt,font=small]{subcaption}
\usepackage[skip=3pt,font=small]{caption}
\usepackage[dvipsnames]{xcolor}
\usepackage[capitalise]{cleveref}
\usepackage{tabularx,colortbl,multirow,array,makecell}
\usepackage{overpic}
\usepackage{tikz}
\usepackage[numbers]{natbib}
\usepackage{multicol}
\usepackage{times}
\usepackage[dvipsnames]{xcolor}
\usepackage{soul}
\usepackage[]{mdframed}
\usepackage{listings}
\usepackage{tabularx,colortbl,multirow,array,makecell}

\definecolor{orange}{RGB}{255, 147, 30}
\definecolor{blue}{RGB}{63, 167, 243}
\definecolor{darkblue}{RGB}{38, 99, 145}
\definecolor{grey}{RGB}{202, 202, 202}
\definecolor{green}{RGB}{34, 139, 34}

\colorlet{lightorange}{orange!20}
\newcommand{\hllo}[1]{%
    {%
    \sethlcolor{lightorange}%
    \hl{#1}%
    }%
}
\colorlet{lightblue}{blue!20}
\newcommand{\hllb}[1]{%
    {%
    \sethlcolor{lightblue}%
    \hl{#1}%
    }%
}
\colorlet{lightgrey}{grey!40}

\definecolor{Gray}{gray}{0.9}
\newcolumntype{x}{>{\columncolor{Gray}}c}
\definecolor{Gray2}{gray}{0.8}
\newcolumntype{y}{>{\columncolor{Gray2}}c}

\newcommand{\framedtext}[1]{%
\par\vspace{1mm} 
\noindent\fbox{%
    \parbox{\dimexpr\linewidth-2\fboxsep-2\fboxrule}{#1}%
}%
\par\vspace{2mm} 
}

\lstdefinelanguage{Python}{
morekeywords={def, for, while, if, elif, else, not, and, or, print, return, from, import},
morecomment=[l]{\#},
morestring=[b]",
morestring=[b]',
}

\lstdefinestyle{PythonStyle}{
    language=Python,
    basicstyle=\ttfamily,
    keywordstyle=\color{darkblue},
    commentstyle=\color{green},
    stringstyle=\color{orange},
    showstringspaces=false,
    breaklines=true,
    breakatwhitespace=true,
    breakindent=0pt, 
    xleftmargin=0pt, 
    xrightmargin=0pt, 
    frame=none, 
    framesep=0pt, 
    aboveskip=0pt, 
    belowskip=0pt, 
}

\newmdenv[
    roundcorner=4pt,
    innertopmargin=3pt,
    innerbottommargin=3pt,
    innerrightmargin=0pt,
    innerleftmargin=3pt,
]{codeframe}

\lstdefinestyle{RawText}{
    basicstyle=\ttfamily\small,
    showstringspaces=false,
    breaklines=true,
    breakatwhitespace=true,
    breakindent=0pt, 
    xleftmargin=0pt, 
    xrightmargin=0pt, 
    frame=none, 
    framesep=0pt, 
    aboveskip=0pt, 
    belowskip=0pt, 
}

\makeatletter
\DeclareRobustCommand\onedot{\futurelet\@let@token\@onedot}
\def\@onedot{\ifx\@let@token.\else.\null\fi\xspace}
\def\eg{\emph{e.g}\onedot} 
\def\ie{\emph{i.e}\onedot} 
 
\def\etc{\emph{etc}\onedot}

\makeatother

\makeatletter
\def\BState{\State\hskip-\ALG@thistlm}
\makeatother

\makeatletter
\renewcommand{\paragraph}{%
  \@startsection{paragraph}{4}%
  {\z@}{0ex \@plus 0ex \@minus 0ex}{-1em}%
  {\hskip\parindent\normalfont\normalsize\bfseries}%
}
\makeatother

\crefname{algocf}{Alg.}{Algs.}
\Crefname{algocf}{Algorithm}{Algorithms}

\definecolor{ggreen}{HTML}{0F9D58}
\definecolor{gblue}{HTML}{4285F4}
\definecolor{gred}{HTML}{DB4437}

\acrodef{tamp}[TAMP]{Task and Motion Planning}
\acrodef{mcts}[MCTS]{Monte Carlo Tree Search}
\acrodef{vln}[VLN]{Vision-Language Navigation}
\acrodef{bc}[BC]{Behavioral Cloning}
\acrodef{iou}[IoU]{Intersection over Union}
\acrodef{hr}[HR]{Human Rating}
\acrodef{sdf}[SDF]{Signed Distance Field}
\acrodef{mlp}[MLP]{Multi-layer Perceptron}
\acrodef{map}[MAP]{Maximum a Posteriori}
\acrodef{gmm}[GMM]{Gaussian Mixture Model}
\acrodef{pddl}[PDDL]{Planning Domain Definition Language}
\acrodef{em}[EM]{Expectation-maximization}
\acrodef{llm}[LLM]{Large Language Model}
\acrodef{vlm}[VLM]{Vision Language Model}
\acrodef{cot}[CoT]{Chain-of-Thought}

\newcommand{\interpret}{InterPreT\xspace}

\SetKwComment{Comment}{//}{}

%% file: main.bbl
\begin{thebibliography}{10}

\bibitem{zhu2021hierarchical}
Yifeng Zhu, Jonathan Tremblay, Stan Birchfield, and Yuke Zhu.
\newblock Hierarchical planning for long-horizon manipulation with geometric and symbolic scene graphs.
\newblock In {\em International Conference on Robotics and Automation (ICRA)}, pages 6541--6548. IEEE, 2021.

\bibitem{xu2019regression}
Danfei Xu, Roberto Mart{\'\i}n-Mart{\'\i}n, De-An Huang, Yuke Zhu, Silvio Savarese, and Li~F Fei-Fei.
\newblock Regression planning networks.
\newblock {\em Advances in Neural Information Processing Systems (NeurIPS)}, 32, 2019.

\bibitem{zhang2023learning}
Zeyu Zhang, Muzhi Han, Baoxiong Jia, Ziyuan Jiao, Yixin Zhu, Song-Chun Zhu, and Hangxin Liu.
\newblock Learning a causal transition model for object cutting.
\newblock In {\em International Conference on Intelligent Robots and Systems (IROS)}, pages 1996--2003. IEEE, 2023.

\bibitem{li2022pre}
Shuang Li, Xavier Puig, Chris Paxton, Yilun Du, Clinton Wang, Linxi Fan, Tao Chen, De-An Huang, Ekin Aky{\"u}rek, Anima Anandkumar, et~al.
\newblock Pre-trained language models for interactive decision-making.
\newblock {\em Advances in Neural Information Processing Systems (NeurIPS)}, 35:31199--31212, 2022.

\bibitem{huang2022language}
Wenlong Huang, Pieter Abbeel, Deepak Pathak, and Igor Mordatch.
\newblock Language models as zero-shot planners: Extracting actionable knowledge for embodied agents.
\newblock In {\em International Conference on Machine Learning (ICML)}, pages 9118--9147. PMLR, 2022.

\bibitem{dong2022survey}
Qingxiu Dong, Lei Li, Damai Dai, Ce~Zheng, Zhiyong Wu, Baobao Chang, Xu~Sun, Jingjing Xu, and Zhifang Sui.
\newblock A survey for in-context learning.
\newblock {\em arXiv preprint arXiv:2301.00234}, 2022.

\bibitem{ahn2022can}
Michael Ahn, Anthony Brohan, Noah Brown, Yevgen Chebotar, Omar Cortes, Byron David, Chelsea Finn, Chuyuan Fu, Keerthana Gopalakrishnan, Karol Hausman, et~al.
\newblock Do as i can, not as i say: Grounding language in robotic affordances.
\newblock {\em arXiv preprint arXiv:2204.01691}, 2022.

\bibitem{huang2022inner}
Wenlong Huang, Fei Xia, Ted Xiao, Harris Chan, Jacky Liang, Pete Florence, Andy Zeng, Jonathan Tompson, Igor Mordatch, Yevgen Chebotar, et~al.
\newblock Inner monologue: Embodied reasoning through planning with language models.
\newblock {\em arXiv preprint arXiv:2207.05608}, 2022.

\bibitem{singh2023progprompt}
Ishika Singh, Valts Blukis, Arsalan Mousavian, Ankit Goyal, Danfei Xu, Jonathan Tremblay, Dieter Fox, Jesse Thomason, and Animesh Garg.
\newblock Progprompt: Generating situated robot task plans using large language models.
\newblock In {\em International Conference on Robotics and Automation (ICRA)}, pages 11523--11530. IEEE, 2023.

\bibitem{liang2023code}
Jacky Liang, Wenlong Huang, Fei Xia, Peng Xu, Karol Hausman, Brian Ichter, Pete Florence, and Andy Zeng.
\newblock Code as policies: Language model programs for embodied control.
\newblock In {\em International Conference on Robotics and Automation (ICRA)}, pages 9493--9500. IEEE, 2023.

\bibitem{liu2023llm+}
Bo~Liu, Yuqian Jiang, Xiaohan Zhang, Qiang Liu, Shiqi Zhang, Joydeep Biswas, and Peter Stone.
\newblock Llm+ p: Empowering large language models with optimal planning proficiency.
\newblock {\em arXiv preprint arXiv:2304.11477}, 2023.

\bibitem{valmeekam2022large}
Karthik Valmeekam, Alberto Olmo, Sarath Sreedharan, and Subbarao Kambhampati.
\newblock Large language models still can't plan (a benchmark for llms on planning and reasoning about change).
\newblock {\em arXiv preprint arXiv:2206.10498}, 2022.

\bibitem{silver2022pddl}
Tom Silver, Varun Hariprasad, Reece~S Shuttleworth, Nishanth Kumar, Tom{\'a}s Lozano-P{\'e}rez, and Leslie~Pack Kaelbling.
\newblock Pddl planning with pretrained large language models.
\newblock In {\em NeurIPS 2022 Foundation Models for Decision Making Workshop}, 2022.

\bibitem{lavalle2006planning}
Steven~M LaValle.
\newblock {\em Planning algorithms}.
\newblock Cambridge university press, 2006.

\bibitem{russell2010artificial}
Stuart~J Russell.
\newblock {\em Artificial intelligence a modern approach}.
\newblock Pearson Education, Inc., 2010.

\bibitem{fikes1971strips}
Richard~E Fikes and Nils~J Nilsson.
\newblock Strips: A new approach to the application of theorem proving to problem solving.
\newblock {\em Artificial intelligence}, 2(3-4):189--208, 1971.

\bibitem{fox2003pddl2}
Maria Fox and Derek Long.
\newblock Pddl2. 1: An extension to pddl for expressing temporal planning domains.
\newblock {\em Journal of Artificial Intelligence Research}, 20:61--124, 2003.

\bibitem{helmert2006fast}
Malte Helmert.
\newblock The fast downward planning system.
\newblock {\em Journal of Artificial Intelligence Research}, 26:191--246, 2006.

\bibitem{kaelbling2011hierarchical}
Leslie~Pack Kaelbling and Tom{\'a}s Lozano-P{\'e}rez.
\newblock Hierarchical task and motion planning in the now.
\newblock In {\em International Conference on Robotics and Automation (ICRA)}, pages 1470--1477. IEEE, 2011.

\bibitem{toussaint2015logic}
Marc Toussaint.
\newblock Logic-geometric programming: An optimization-based approach to combined task and motion planning.
\newblock In {\em International Joint Conference on Artificial Intelligence (IJCAI)}, pages 1930--1936, 2015.

\bibitem{garrett2020pddlstream}
Caelan~Reed Garrett, Tom{\'a}s Lozano-P{\'e}rez, and Leslie~Pack Kaelbling.
\newblock Pddlstream: Integrating symbolic planners and blackbox samplers via optimistic adaptive planning.
\newblock In {\em Proceedings of the International Conference on Automated Planning and Scheduling}, volume~30, pages 440--448, 2020.

\bibitem{konidaris2018skills}
George Konidaris, Leslie~Pack Kaelbling, and Tomas Lozano-Perez.
\newblock From skills to symbols: Learning symbolic representations for abstract high-level planning.
\newblock {\em Journal of Machine Learning Research (JMLR)}, 61:215--289, 2018.

\bibitem{james2022autonomous}
Steven James, Benjamin Rosman, and George Konidaris.
\newblock Autonomous learning of object-centric abstractions for high-level planning.
\newblock In {\em International Conference on Learning Representations (ICLR)}, 2021.

\bibitem{loula2019discovering}
Jo{\~a}o Loula, Tom Silver, Kelsey~R Allen, and Josh Tenenbaum.
\newblock Discovering a symbolic planning language from continuous experience.
\newblock In {\em Annual Meeting of the Cognitive Science Society (CogSci)}, page 2193, 2019.

\bibitem{silver2023predicate}
Tom Silver, Rohan Chitnis, Nishanth Kumar, Willie McClinton, Tom{\'a}s Lozano-P{\'e}rez, Leslie Kaelbling, and Joshua~B Tenenbaum.
\newblock Predicate invention for bilevel planning.
\newblock In {\em AAAI Conference on Artificial Intelligence (AAAI)}, volume~37, pages 12120--12129, 2023.

\bibitem{ahmetoglu2022deepsym}
Alper Ahmetoglu, M~Yunus Seker, Justus Piater, Erhan Oztop, and Emre Ugur.
\newblock Deepsym: Deep symbol generation and rule learning for planning from unsupervised robot interaction.
\newblock {\em Journal of Artificial Intelligence Research}, 75:709--745, 2022.

\bibitem{mandler1992build}
Jean~M Mandler.
\newblock How to build a baby: Ii. conceptual primitives.
\newblock {\em Psychological review}, 99(4):587, 1992.

\bibitem{goksun2010trading}
Tilbe G{\"o}ksun, Kathy Hirsh-Pasek, and Roberta Michnick~Golinkoff.
\newblock Trading spaces: Carving up events for learning language.
\newblock {\em Perspectives on Psychological Science}, 5(1):33--42, 2010.

\bibitem{ma2023eureka}
Yecheng~Jason Ma, William Liang, Guanzhi Wang, De-An Huang, Osbert Bastani, Dinesh Jayaraman, Yuke Zhu, Linxi Fan, and Anima Anandkumar.
\newblock Eureka: Human-level reward design via coding large language models.
\newblock {\em arXiv preprint arXiv:2310.12931}, 2023.

\bibitem{liu2023interactive}
Huihan Liu, Alice Chen, Yuke Zhu, Adith Swaminathan, Andrey Kolobov, and Ching-An Cheng.
\newblock Interactive robot learning from verbal correction.
\newblock {\em arXiv preprint arXiv:2310.17555}, 2023.

\bibitem{zhao2023survey}
Wayne~Xin Zhao, Kun Zhou, Junyi Li, Tianyi Tang, Xiaolei Wang, Yupeng Hou, Yingqian Min, Beichen Zhang, Junjie Zhang, Zican Dong, et~al.
\newblock A survey of large language models.
\newblock {\em arXiv preprint arXiv:2303.18223}, 2023.

\bibitem{openai2023gpt4}
OpenAI.
\newblock Gpt-4 technical report.
\newblock {\em arXiv preprint arXiv:2303.08774}, 2023.

\bibitem{kojima2022large}
Takeshi Kojima, Shixiang~Shane Gu, Machel Reid, Yutaka Matsuo, and Yusuke Iwasawa.
\newblock Large language models are zero-shot reasoners.
\newblock {\em Advances in Neural Information Processing Systems (NeurIPS)}, 35:22199--22213, 2022.

\bibitem{yao2022react}
Shunyu Yao, Jeffrey Zhao, Dian Yu, Nan Du, Izhak Shafran, Karthik~R Narasimhan, and Yuan Cao.
\newblock React: Synergizing reasoning and acting in language models.
\newblock In {\em International Conference on Learning Representations (ICLR)}, 2022.

\bibitem{chen2021evaluating}
Mark Chen, Jerry Tworek, Heewoo Jun, Qiming Yuan, Henrique Ponde de~Oliveira Pinto, Jared Kaplan, Harri Edwards, Yuri Burda, Nicholas Joseph, Greg Brockman, et~al.
\newblock Evaluating large language models trained on code.
\newblock {\em arXiv preprint arXiv:2107.03374}, 2021.

\bibitem{silver2021learning}
Tom Silver, Rohan Chitnis, Joshua Tenenbaum, Leslie~Pack Kaelbling, and Tom{\'a}s Lozano-P{\'e}rez.
\newblock Learning symbolic operators for task and motion planning.
\newblock In {\em International Conference on Intelligent Robots and Systems (IROS)}, pages 3182--3189. IEEE, 2021.

\bibitem{pasula2007learning}
Hanna~M Pasula, Luke~S Zettlemoyer, and Leslie~Pack Kaelbling.
\newblock Learning symbolic models of stochastic domains.
\newblock {\em Journal of Artificial Intelligence Research}, 29:309--352, 2007.

\bibitem{jetchev2013learning}
Nikolay Jetchev, Tobias Lang, and Marc Toussaint.
\newblock Learning grounded relational symbols from continuous data for abstract reasoning.
\newblock In {\em Proceedings of the 2013 ICRA Workshop on Autonomous Learning}, 2013.

\bibitem{ugur2015bottom}
Emre Ugur and Justus Piater.
\newblock Bottom-up learning of object categories, action effects and logical rules: From continuous manipulative exploration to symbolic planning.
\newblock In {\em International Conference on Robotics and Automation (ICRA)}, pages 2627--2633. IEEE, 2015.

\bibitem{rosen2023synthesizing}
Eric Rosen, Steven James, Sergio Orozco, Vedant Gupta, Max Merlin, Stefanie Tellex, and George Konidaris.
\newblock Synthesizing navigation abstractions for planning with portable manipulation skills.
\newblock In {\em Conference on Robot Learning (CoRL)}, pages 2278--2287. PMLR, 2023.

\bibitem{curtis2022discovering}
Aidan Curtis, Tom Silver, Joshua~B Tenenbaum, Tom{\'a}s Lozano-P{\'e}rez, and Leslie Kaelbling.
\newblock Discovering state and action abstractions for generalized task and motion planning.
\newblock In {\em AAAI Conference on Artificial Intelligence (AAAI)}, volume~36, pages 5377--5384, 2022.

\bibitem{umili2021learning}
Elena Umili, Emanuele Antonioni, Francesco Riccio, Roberto Capobianco, Daniele Nardi, and Giuseppe De~Giacomo.
\newblock Learning a symbolic planning domain through the interaction with continuous environments.
\newblock In {\em ICAPS PRL Workshop}, 2021.

\bibitem{asai2019unsupervised}
Masataro Asai.
\newblock Unsupervised grounding of plannable first-order logic representation from images.
\newblock In {\em Proceedings of the International Conference on Automated Planning and Scheduling}, volume~29, pages 583--591, 2019.

\bibitem{xu2017scene}
Danfei Xu, Yuke Zhu, Christopher~B Choy, and Li~Fei-Fei.
\newblock Scene graph generation by iterative message passing.
\newblock In {\em Conference on Computer Vision and Pattern Recognition (CVPR)}, pages 5410--5419, 2017.

\bibitem{mao2019neuro}
Jiayuan Mao, Chuang Gan, Pushmeet Kohli, Joshua~B Tenenbaum, and Jiajun Wu.
\newblock The neuro-symbolic concept learner: Interpreting scenes, words, and sentences from natural supervision.
\newblock {\em arXiv preprint arXiv:1904.12584}, 2019.

\bibitem{krishna2017visual}
Ranjay Krishna, Yuke Zhu, Oliver Groth, Justin Johnson, Kenji Hata, Joshua Kravitz, Stephanie Chen, Yannis Kalantidis, Li-Jia Li, David~A Shamma, et~al.
\newblock Visual genome: Connecting language and vision using crowdsourced dense image annotations.
\newblock {\em International Journal of Computer Vision (IJCV)}, 123:32--73, 2017.

\bibitem{zhang2023grounding}
Xiaohan Zhang, Yan Ding, Saeid Amiri, Hao Yang, Andy Kaminski, Chad Esselink, and Shiqi Zhang.
\newblock Grounding classical task planners via vision-language models.
\newblock {\em arXiv preprint arXiv:2304.08587}, 2023.

\bibitem{kamath2023s}
Amita Kamath, Jack Hessel, and Kai-Wei Chang.
\newblock What's" up" with vision-language models? investigating their struggle with spatial reasoning.
\newblock {\em arXiv preprint arXiv:2310.19785}, 2023.

\bibitem{guo2023doremi}
Yanjiang Guo, Yen-Jen Wang, Lihan Zha, Zheyuan Jiang, and Jianyu Chen.
\newblock Doremi: Grounding language model by detecting and recovering from plan-execution misalignment.
\newblock {\em arXiv preprint arXiv:2307.00329}, 2023.

\bibitem{bobu2022learning}
Andreea Bobu, Chris Paxton, Wei Yang, Balakumar Sundaralingam, Yu-Wei Chao, Maya Cakmak, and Dieter Fox.
\newblock Learning perceptual concepts by bootstrapping from human queries.
\newblock {\em IEEE Robotics and Automation Letters (RA-L)}, 7(4):11260--11267, 2022.

\bibitem{lis2023embodied}
Amber Li and Tom Silver.
\newblock Embodied active learning of relational state abstractions for bilevel planning.
\newblock In {\em Conference on Lifelong Learning Agents (CoLLAs)}, 2023.

\bibitem{migimatsu2022grounding}
Toki Migimatsu and Jeannette Bohg.
\newblock Grounding predicates through actions.
\newblock In {\em International Conference on Robotics and Automation (ICRA)}, pages 3498--3504. IEEE, 2022.

\bibitem{mao2022pdsketch}
Jiayuan Mao, Tom{\'a}s Lozano-P{\'e}rez, Josh Tenenbaum, and Leslie Kaelbling.
\newblock Pdsketch: Integrated domain programming, learning, and planning.
\newblock {\em Advances in Neural Information Processing Systems (NeurIPS)}, 35:36972--36984, 2022.

\bibitem{brown2020language}
Tom Brown, Benjamin Mann, Nick Ryder, Melanie Subbiah, Jared~D Kaplan, Prafulla Dhariwal, Arvind Neelakantan, Pranav Shyam, Girish Sastry, Amanda Askell, et~al.
\newblock Language models are few-shot learners.
\newblock {\em Advances in Neural Information Processing Systems (NeurIPS)}, 33:1877--1901, 2020.

\bibitem{shinn2023reflexion}
Noah Shinn, Federico Cassano, Edward Berman, Ashwin Gopinath, Karthik Narasimhan, and Shunyu Yao.
\newblock Reflexion: Language agents with verbal reinforcement learning, 2023.

\bibitem{wang2023jarvis}
Zihao Wang, Shaofei Cai, Anji Liu, Yonggang Jin, Jinbing Hou, Bowei Zhang, Haowei Lin, Zhaofeng He, Zilong Zheng, Yaodong Yang, et~al.
\newblock Jarvis-1: Open-world multi-task agents with memory-augmented multimodal language models.
\newblock {\em arXiv preprint arXiv:2311.05997}, 2023.

\bibitem{song2023llm}
Chan~Hee Song, Jiaman Wu, Clayton Washington, Brian~M Sadler, Wei-Lun Chao, and Yu~Su.
\newblock Llm-planner: Few-shot grounded planning for embodied agents with large language models.
\newblock In {\em International Conference on Computer Vision (ICCV)}, pages 2998--3009, 2023.

\bibitem{wu2023embodied}
Zhenyu Wu, Ziwei Wang, Xiuwei Xu, Jiwen Lu, and Haibin Yan.
\newblock Embodied task planning with large language models.
\newblock {\em arXiv preprint arXiv:2307.01848}, 2023.

\bibitem{du2023vision}
Yuqing Du, Ksenia Konyushkova, Misha Denil, Akhil Raju, Jessica Landon, Felix Hill, Nando de~Freitas, and Serkan Cabi.
\newblock Vision-language models as success detectors.
\newblock {\em arXiv preprint arXiv:2303.07280}, 2023.

\bibitem{wang2024llm3}
Shu Wang, Muzhi Han, Ziyuan Jiao, Zeyu Zhang, Ying~Nian Wu, Song-Chun Zhu, and Hangxin Liu.
\newblock Llm3:large language model-based task and motion planning with motion failure reasoning.
\newblock {\em arXiv preprint arXiv:2403.11552}, 2024.

\bibitem{zhi2024closed}
Peiyuan Zhi, Zhiyuan Zhang, Muzhi Han, Zeyu Zhang, Zhitian Li, Ziyuan Jiao, Baoxiong Jia, and Siyuan Huang.
\newblock Closed-loop open-vocabulary mobile manipulation with gpt-4v.
\newblock {\em arXiv preprint arXiv:2404.10220}, 2024.

\bibitem{xie2023translating}
Yaqi Xie, Chen Yu, Tongyao Zhu, Jinbin Bai, Ze~Gong, and Harold Soh.
\newblock Translating natural language to planning goals with large-language models.
\newblock {\em arXiv preprint arXiv:2302.05128}, 2023.

\bibitem{yu2023language}
Wenhao Yu, Nimrod Gileadi, Chuyuan Fu, Sean Kirmani, Kuang-Huei Lee, Montse~Gonzalez Arenas, Hao-Tien~Lewis Chiang, Tom Erez, Leonard Hasenclever, Jan Humplik, et~al.
\newblock Language to rewards for robotic skill synthesis.
\newblock {\em arXiv preprint arXiv:2306.08647}, 2023.

\bibitem{wang2023voyager}
Guanzhi Wang, Yuqi Xie, Yunfan Jiang, Ajay Mandlekar, Chaowei Xiao, Yuke Zhu, Linxi Fan, and Anima Anandkumar.
\newblock Voyager: An open-ended embodied agent with large language models.
\newblock {\em arXiv preprint arXiv:2305.16291}, 2023.

\bibitem{wei2022chain}
Jason Wei, Xuezhi Wang, Dale Schuurmans, Maarten Bosma, Fei Xia, Ed~Chi, Quoc~V Le, Denny Zhou, et~al.
\newblock Chain-of-thought prompting elicits reasoning in large language models.
\newblock {\em Advances in Neural Information Processing Systems (NeurIPS)}, 35:24824--24837, 2022.

\bibitem{wang2018active}
Zi~Wang, Caelan~Reed Garrett, Leslie~Pack Kaelbling, and Tom{\'a}s Lozano-P{\'e}rez.
\newblock Active model learning and diverse action sampling for task and motion planning.
\newblock In {\em International Conference on Intelligent Robots and Systems (IROS)}, pages 4107--4114. IEEE, 2018.

\bibitem{gupta1991complexity}
Naresh Gupta, Dana~S Nau, et~al.
\newblock Complexity results for blocks-world planning.
\newblock In {\em AAAI Conference on Artificial Intelligence (AAAI)}, volume~91, pages 629--633. Citeseer, 1991.

\bibitem{chat2023openai}
OpenAI.
\newblock Chatgpt.
\newblock \url{https://chat.openai.com/}.

\bibitem{garrett2021integrated}
Caelan~Reed Garrett, Rohan Chitnis, Rachel Holladay, Beomjoon Kim, Tom Silver, Leslie~Pack Kaelbling, and Tom{\'a}s Lozano-P{\'e}rez.
\newblock Integrated task and motion planning.
\newblock {\em Annual review of control, robotics, and autonomous systems}, 4:265--293, 2021.

\bibitem{kumar2023learning}
Nishanth Kumar, Willie McClinton, Rohan Chitnis, Tom Silver, Tom{\'a}s Lozano-P{\'e}rez, and Leslie~Pack Kaelbling.
\newblock Learning efficient abstract planning models that choose what to predict.
\newblock In {\em Conference on Robot Learning (CoRL)}, pages 2070--2095. PMLR, 2023.

\bibitem{ren2024grounded}
Tianhe Ren, Shilong Liu, Ailing Zeng, Jing Lin, Kunchang Li, He~Cao, Jiayu Chen, Xinyu Huang, Yukang Chen, Feng Yan, et~al.
\newblock Grounded sam: Assembling open-world models for diverse visual tasks.
\newblock {\em arXiv preprint arXiv:2401.14159}, 2024.

\bibitem{zhu2022viola}
Yifeng Zhu, Abhishek Joshi, Peter Stone, and Yuke Zhu.
\newblock Viola: Imitation learning for vision-based manipulation with object proposal priors.
\newblock {\em arXiv preprint arXiv:2210.11339}, 2022.

\end{thebibliography}
